\documentclass[letterpaper]{article}
\usepackage[preprint]{aaai2027}
\usepackage[hyphens]{url}
\usepackage{graphicx}
\usepackage{natbib}
\usepackage{caption}
\usepackage{booktabs}
\usepackage{xcolor}

\usepackage{float}
\usepackage{dblfloatfix}
\floatstyle{ruled}
\newfloat{algorithm}{t}{alg}
\floatname{algorithm}{Algorithm}
\usepackage{dcolumn}
\usepackage{amsmath}
\usepackage{amssymb}
\newcolumntype{L}[1]{>{\raggedright\arraybackslash}m{#1}}
\newcolumntype{C}[1]{>{\centering\arraybackslash}m{#1}}
\newcolumntype{d}{D{.}{.}{-1}}
\title{When Model Priors Conflict with Visual Evidence: Mitigating Commonsense-Driven Hallucinations by Selective Prior Calibration}
\author{Kesheng Chen, Yamin Hu, Wenjian Luo}
\affiliations{
Guangdong Provincial Key Laboratory of Novel Security Intelligence Technologies\\
Institute of Cyberspace Security\\
School of Computer Science and Technology, Harbin Institute of Technology, Shenzhen, China\\
22s151138@stu.hit.edu.cn, huyamin@hit.edu.cn, luowenjian@hit.edu.cn
}

\begin{document}
\maketitle

\begin{abstract}
In vision--language models, commonsense-driven hallucination (CDH) occurs when a model's commonsense prior overrides clear visual evidence of an atypical state. For example, a model may report that a visibly six-fingered hand has five fingers. We show that these errors are systematically directed: when a model answers a question about a counterfactual (CF) image incorrectly, its answer often coincides with the candidate it prefers without access to the image. Suppressing this prior indiscriminately can repair CF errors, but may also disrupt correct answers on matched commonsense (CS) images, where the same prior is helpful. We therefore propose Selective Prior Calibration (SPC), which subtracts candidate-level prior-preference estimates from image-conditioned scores with an instance-dependent strength and revises the original prediction only when the resulting score pattern strongly supports an alternative. Extensive experiments demonstrate that SPC substantially improves accuracy on CF images while largely preserving accuracy on matched CS images. Furthermore, these gains generalize across CDH categories, candidate-answer permutations, and other conflict benchmarks, while SPC rarely alters predictions on benchmarks without such conflicts.

\end{abstract}

\begin{center}
\small\textbf{Code:} \url{https://github.com/MiLab-HITSZ/2026ChenSPC}
\end{center}

\section{Introduction}

Commonsense-driven hallucination (CDH) \cite{chen2026cdh} occurs when a vision--language model's commonsense prior overrides clear visual evidence of an atypical state. For example, consider a commonsense (CS) image of a one-headed dog and a counterfactual (CF) image of a visibly two-headed dog, both queried with the same question: ``How many heads does the dog have?'' On the CS image, visual evidence and the model's prior both favor one, and the model answers correctly. On the CF image, however, the model again answers one, exemplifying CDH: visual evidence supports two, whereas the model's prior favors the ordinary answer one.  Such atypical states also occur in real-world images, including those depicting rare anatomical variations \cite{chen2026cdh}. Consequently, mitigating CDH is essential for ensuring that models correctly interpret unusual but valid visual evidence.

CDH is related to, but distinct from, object hallucination. Object-hallucination benchmarks evaluate unsupported objects, attributes, and relations \cite{rohrbach2018chair,li2023pope,wang2023amber,sun2024mmhal,zheng2025reefknot}. Existing object-hallucination mitigation methods modify token decoding \cite{leng2024vcd,favero2024m3id,huang2024opera,chen2024halc}, strengthen attention to visual evidence \cite{liu2024pai}, or steer latent representations \cite{wu2026revis}. More directly related to CDH, prior studies examine language bias and conflicts between visual evidence and world knowledge \cite{goyal2017vqa,agrawal2018priors,cadene2019rubi,chen2020css,niu2021cfvqa,selvaraju2019hint,zhou2023rome,lee2025vlind,zhou2025causalmm,liu2025insight,golovanevsky2025pixels}. Most closely related to our work, NoLan contrasts multimodal and text-only token distributions to suppress language priors in object hallucination \cite{ren2026nolan}. Taken together, existing work leaves two questions unresolved: whether erroneous responses systematically converge on the ordinary answer favored by the model's prior, and how this error direction can be mitigated without harming cases in which that answer is visually correct.

To investigate these questions, we use CDH-Bench \cite{chen2026cdh}, which consists of matched CF--CS image pairs. The two images in each pair share the same question and candidate answer set. The CF image depicts a state that conflicts with ordinary-world expectations, whereas the matched CS image depicts the corresponding ordinary state. To estimate the model's prior preference, we score the candidate answers without the image and identify the highest-scoring candidate. We find that incorrect predictions on CF images systematically converge on this prior-favored candidate, a pattern we call \emph{directed prior attraction}. Notably, this candidate frequently matches the ground-truth answer for the paired CS image, indicating that the errors are generally directed toward the answer associated with the ordinary state.

The observed direction of CF errors suggests a simple intervention: subtracting prior preferences from image-conditioned candidate scores. Applying this correction indiscriminately, however, can harm performance on CS images, where the prior reinforces the answer indicated by visual evidence. Effective CDH mitigation must therefore balance CF repair against CS retention.  We address this trade-off by proposing \emph{Selective Prior Calibration} (SPC), which learns an instance-dependent strength for subtracting candidate-level prior-preference estimates and revises the original prediction only when the adjusted score pattern provides sufficient support for an alternative. Notably, SPC processes each image independently during online inference, without access to its matched counterpart or a label indicating whether the current image is CF or CS. To our knowledge, SPC is the first method specifically designed to exploit the answer-level direction of CDH while explicitly balancing CF repair and CS retention.

Our contributions are threefold. First, we show that incorrect CF predictions systematically converge on the prior-favored candidate, revealing directed prior attraction. Second, we propose SPC, which learns how strongly to subtract candidate-level prior-preference estimates from image-conditioned scores and selectively revises the original prediction.  Third, extensive experiments show that SPC substantially improves accuracy on CF images while largely preserving accuracy on matched CS images. These gains generalize across CDH categories, candidate-answer permutations, models, and related conflict benchmarks, while SPC rarely alters predictions on benchmarks without prior--evidence conflict.

\section{Preliminaries}

\subsection{CDH Mitigation Objective}
CDH highlights the dual role of commonsense priors in visual understanding.
In a counterfactual (CF) image, an atypical visual state conflicts with the
model's commonsense prior, so a correct response requires the model to
prioritize visual evidence over the ordinary-state answer favored by that
prior. In a commonsense (CS) image, visual evidence and the prior
agree, allowing the same prior to support a correct prediction.
Indiscriminately weakening the prior may therefore repair CF errors while
introducing new errors on CS images. Effective CDH mitigation must reduce
harmful prior influence under conflict while preserving correct predictions
when the prior agrees with visual evidence.

\subsection{CDH-Bench}

CDH-Bench, constructed by Chen et al.~\cite{chen2026cdh}, provides a controlled setting for studying CDH mitigation. It contains 300 instances spanning 14 subcategories, grouped into three broad categories: counting, attributes, and relations. Each instance consists of a matched CF--CS image pair. Each pair is associated with a question evaluated in two tasks: four-option multiple-choice QA and binary yes/no QA, hereafter denoted MC and QA, respectively.  Let $\mathcal{Y}$ denote the task-specific candidate set: $\{\mathrm{A},\mathrm{B},\mathrm{C},\mathrm{D}\}$ for MC and $\{\text{yes},\text{no}\}$ for QA.

Within each instance, the CF and CS images share the same question and candidate set. Only the depicted state and, consequently, the visually correct answer differ. This pairing enables us to test whether CF errors systematically converge on a prior-favored answer, indicating directed prior attraction. It also allows us to assess whether a mitigation method repairs errors under prior--evidence conflict without disrupting correct predictions when the prior agrees with visual evidence.

\section{Selective Prior Calibration}

\begin{figure*}[!t]
\centering
\includegraphics[width=0.9\textwidth]{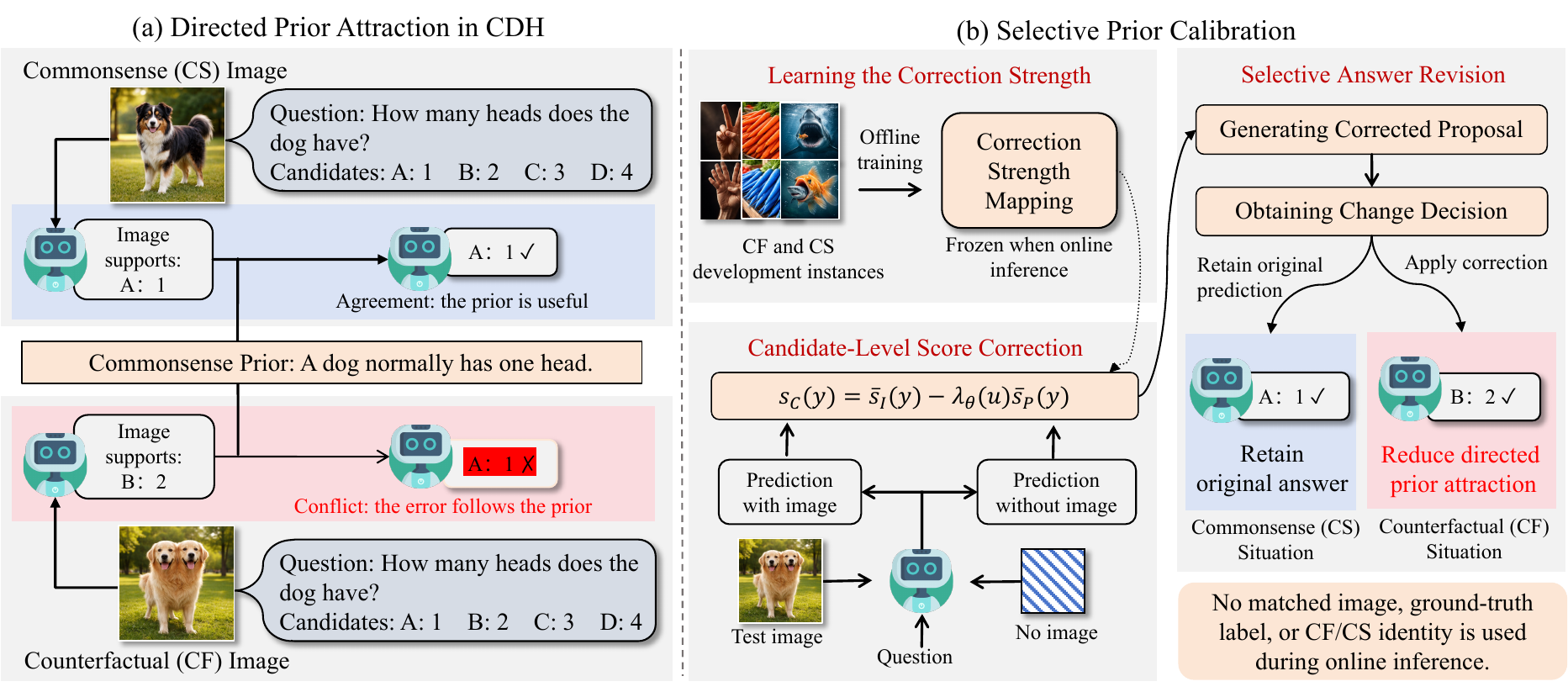}
\caption{\textbf{From directed prior attraction to selective prior calibration.}
(a) On the CS image, visual evidence aligns with the model's prior. On the CF
image, they conflict, and the same prior biases the model toward the
prior-favored answer. (b) SPC scores candidates with and without the image, adjusts their scores using a learned correction strength, and adopts the corrected proposal only when
it is sufficiently supported.}
\label{fig:overview}
\end{figure*}

Directed prior attraction motivates candidate-level correction based on estimated prior preference: when a candidate receives strong estimated prior support, its image-conditioned score is reduced accordingly. Based on this idea, we propose Selective Prior Calibration (SPC) to perform this correction adaptively and selectively. During development, we use paired CF and CS data to fit the parameters that determine the correction strength for each instance. This paired learning promotes CF repair while discouraging unnecessary changes to correct predictions on CS images. During online inference, SPC first scores each candidate with and without the image. It then uses the learned parameters to determine the correction strength, adjust the candidate scores, and form a corrected proposal. Finally, it replaces the original answer only when the proposal satisfies the predefined support conditions. Figure~\ref{fig:overview} illustrates the directed prior attraction and the overall procedure.

\subsection{Candidate-Level Score Correction}

Given an image--question pair and candidate set $\mathcal{Y}$, SPC scores
each candidate $y\in\mathcal{Y}$ both with and without the image, using the
log-probability of its corresponding answer token. Scoring with the image yields the
image-conditioned score $s_I(y)$, whereas repeating the same procedure
without the image yields the text-only score $s_P(y)$. We use $s_P(y)$ as
an operational estimate of the model's prior preference for candidate $y$.
It captures preferences arising from the model's learned knowledge and
should be interpreted as a proxy rather than a direct measure of its latent
prior \cite{favero2024m3id,ren2026nolan}.
We mean-center $s_I$ and $s_P$ separately across candidates, obtaining
$\bar{s}_I$ and $\bar{s}_P$; this preserves candidate rankings and pairwise
score differences while standardizing the score levels used for calibration.

SPC then adjusts each candidate's score according to its estimated prior
preference:
\begin{equation}
s_C(y)=\bar{s}_I(y)-\lambda\bar{s}_P(y),
\qquad y\in\mathcal{Y},
\label{eq:spc}
\end{equation}
where $\lambda\geq0$ controls the correction strength. 
In a CDH case, the incorrect ordinary-state candidate typically receives
strong support from the text-only score $\bar{s}_P$. Equation~(\ref{eq:spc})
therefore subtracts more from this candidate than from alternatives,
weakening its prior-driven advantage. If another candidate retains sufficient
support in the image-conditioned scores, it can become the highest-scoring
candidate after correction. SPC thus mitigates CDH by discounting candidate
support that persists even when the image is removed.

However, the same correction can be harmful on a CS example, where the
prior-favored candidate is also visually correct. To explain
this trade-off, let $y^*$ denote the visually supported answer and $y$ a
competing candidate. Their corrected relative score is
\begin{equation}
\begin{split}
s_C(y^*)-s_C(y)
={}&\bar{s}_I(y^*)-\bar{s}_I(y)\\
&+\lambda\bigl[\bar{s}_P(y)-\bar{s}_P(y^*)\bigr].
\end{split}
\label{eq:relative-score}
\end{equation}
When the prior estimate favors an incorrect ordinary-state candidate $y$,
as often occurs in CF errors, the final term is positive and increases the
score of $y^*$ relative to $y$. When the prior instead favors the visually
correct answer $y^*$, as is common on CS examples, the final term is negative
and reduces its relative score. A strong fixed correction may therefore
repair CF errors while introducing new CS errors. SPC must instead adapt the correction strength to each instance
rather than apply a fixed strength universally.

\subsection{Learning the Correction Strength}

SPC predicts an instance-dependent correction strength from 13 features
computed from the image-conditioned scores, text-only scores, original
prediction, and candidate-set size. These features include distribution
entropy, peak scores, differences between the two highest scores,
Jensen--Shannon divergence, and indicators of agreement among the
predictions. The complete feature definition is provided in the appendix.

We learn the mapping from these
features to the correction strength based on a development set. To support fitting with limited paired development data, we parameterize
this mapping with a low-capacity function:
\begin{equation}
\lambda_\theta(u)=
\min\!\left\{
\operatorname{softplus}(u^\top\theta),
\lambda_{\max}
\right\},
\label{eq:lambda}
\end{equation}
where $u$ is the 13-dimensional feature vector and $\theta$ is the learned
parameter vector. The $\operatorname{softplus}$ function constrains the correction strength
to be positive, while $\lambda_{\max}$ prevents excessively large
adjustments.

We learn $\theta$ by minimizing a regularized cross-entropy loss over the
corrected candidate scores. Specifically, we place an isotropic Gaussian
prior on $\theta$,
$\theta\sim\mathcal{N}(0,\rho^{-1}I)$, and obtain its maximum a posteriori
(MAP) estimate by minimizing
\begin{align}
\ell_i(\theta)
&=
-\log\!\left[
\operatorname{softmax}\!\left(
\bar{s}_{I,i}
-\lambda_\theta(u_i)\bar{s}_{P,i}
\right)
\right]_{y_i},
\nonumber\\
\hat{\theta}
&=
\arg\min_{\theta}
\sum_{i\in D}\ell_i(\theta)
+\frac{\rho}{2}\lVert\theta\rVert_2^2,
\label{eq:fit}
\end{align}
where $D$ denotes the development instances and $y_i$ is the ground-truth
answer for instance $i$.

We fit one shared parameter vector $\theta$ jointly using paired CF and CS
development data from both the MC and QA tasks. Each development instance in CDH-Bench contributes four training instances: one CF and
one CS instance for each task. The CF instances provide supervision for removing harmful prior preference, whereas the CS instances penalize excessive correction when the prior supports the correct answer. The appendix provides the complete fitting details and analyzes when the
learned correction recovers the visually supported answer.
\subsection{Selective Answer Revision}

Equation~(\ref{eq:spc}) with adaptive correction strength always produces a highest-scoring candidate, but this
candidate is not necessarily a reliable correction. For example, subtraction may reverse the order of two nearly tied candidates, causing an alternative to rank first by only a small amount. These corrections are particularly risky when
the original answer is already correct, as is often the case for CS images.
SPC therefore treats the highest-scoring candidate after correction as a
proposal rather than an automatic replacement. It changes the original answer
only when the proposal is sufficiently separated from the remaining
candidates and the score pattern is consistent with one of the predefined
support conditions.

To define the change rule, we introduce the following notation. Let
$i_b\in\mathcal{Y}$ denote the model's original prediction, obtained by
parsing its standard response to the image--question prompt containing the
candidate set $\mathcal{Y}$. We denote the highest-scoring candidates under
image-conditioned and text-only scoring by
\begin{equation}
i_I=\arg\max_{y\in\mathcal{Y}}\bar{s}_I(y),
\qquad
i_P=\arg\max_{y\in\mathcal{Y}}\bar{s}_P(y),
\end{equation}
respectively. Here, $i_P$ is the prior-favored candidate and often
corresponds to the ordinary-state answer in the CDH cases.
The corrected candidate probabilities and the resulting proposal are
\begin{equation}
q_C=\operatorname{softmax}(s_C),
\qquad
z_M=\arg\max_{y\in\mathcal{Y}}q_C(y),
\end{equation}
where $z_M$ denotes the proposed corrected answer. We quantify its support
by its probability gap over the strongest alternative:
\begin{equation}
m=q_C(z_M)-\max_{y\in\mathcal{Y},\,y\ne z_M}q_C(y).
\end{equation}

SPC restricts answer changes to two predefined support conditions. The first
is $ d=(i_I\ne i_P)\land(z_M=i_I) $, where the image-conditioned and text-only scores favor different candidates, and the corrected scores select the image-conditioned candidate. The second is $ a=(i_b=i_I=i_P)\land(z_M\ne i_b) $, where the original prediction and both scoring conditions initially favor the same candidate, but prior subtraction promotes an alternative.

The corrected proposal is adopted only if it differs from the original
answer, has sufficient support, uses a candidate-set size observed during
development, and satisfies either support condition. Define the Boolean
change decision as
\begin{equation}
g_M={}(z_M\ne i_b)\land(m\ge m_{\min})\land(K\in\mathcal{K}_D)\land(d\lor a).
\end{equation}
The final answer is
\begin{equation}
\hat{y}=
\begin{cases}
z_M, & \text{if } g_M,\\
i_b, & \text{otherwise}.
\end{cases}
\label{eq:gate}
\end{equation}
Here, $m_{\min}$ is selected on the development set, and
$\mathcal{K}_D$ contains the candidate-set sizes observed during
development. 

Notably, during online inference, the change rule uses only the input image, question, candidate set, original prediction, and the corresponding image-conditioned and text-only scores. It does not require the matched image, ground-truth answer, or information about whether the input image is a CF or CS image.

\section{Experimental Setup}

We organize our experiments around five research questions. \textbf{RQ1: Do CDH errors systematically converge on a prior-favored candidate?} We analyze the direction of incorrect CF predictions.  \textbf{RQ2: How effectively does SPC repair CF errors while preserving correct predictions on matched CS images?} We evaluate CF repair, CS retention, and performance relative to existing methods.  \textbf{RQ3: Which components drive SPC's performance?} We examine candidate-specific prior subtraction, instance-dependent correction strength, and selective answer revision.  \textbf{RQ4: How well does SPC generalize beyond its fitting setting?} We test held-out CDH categories, unseen candidate-answer orders, and external benchmarks.
\textbf{RQ5: Under what conditions is prior-based correction effective?} We
examine its sensitivity to question form and claim order.

\paragraph{Benchmarks and models.}
We reserve five instances from each of the 14 subcategories of CDH-Bench \cite{chen2026cdh} to form a development set of 70 instances. The remaining 230 instances form the test set. We evaluate Qwen3-VL-32B \cite{bai2025qwen3vl} and LLaVA-1.6-34B \cite{liu2024llavabaselines} on both MC and QA tasks. SPC requires a fixed candidate set and does not currently support open-ended generation.

We evaluate generalization on Visual CounterFact \cite{golovanevsky2025pixels}, HallusionBench \cite{guan2024hallusionbench}, ConflictVIS \cite{liu2025insight}, POPE \cite{li2023pope}, and POPEv2 \cite{li2026popev2}.

\paragraph{Evaluation metrics.}
To evaluate both CF repair and CS retention, let $b$ denote the original model
and $m$ a mitigation method. We compute
$\Delta_{\mathrm{CF}}=Acc^{m}_{\mathrm{CF}}-Acc^{b}_{\mathrm{CF}}$ and
$\Delta_{\mathrm{CS}}=Acc^{m}_{\mathrm{CS}}-Acc^{b}_{\mathrm{CS}}$.
Here, $\Delta_{\mathrm{CF}}$ measures whether mitigation recovers answers supported by
atypical visual evidence, whereas $\Delta_{\mathrm{CS}}$ measures whether it preserves
accuracy when visual evidence agrees with the commonsense prior. A useful
method should increase $\Delta_{\mathrm{CF}}$ while keeping $\Delta_{\mathrm{CS}}$ close to zero
or positive. We also report the number of repairs and harms, which count corrected errors
and newly introduced errors, respectively.  We additionally report absolute CF and CS accuracy (CF-Acc and CS-Acc). Following CDH-Bench \cite{chen2026cdh}, we also report CFAD,
RPD, and CCR. We assess paired accuracy changes using exact
McNemar tests and estimate their variability using 10,000 pair-level bootstrap
resamples.

\paragraph{Baselines and model selection.}
We compare Focus on Vision \cite{liu2025insight}, VCD \cite{leng2024vcd}, MFCD \cite{liu2025mfcd}, PAI \cite{liu2024pai}, NoLan \cite{ren2026nolan}, and REVIS \cite{wu2026revis}. We retain their released interventions and settings. Implementation details are
in the appendix.

We select the hyperparameter configurations of SPC by
maximizing $\Delta_{\mathrm{CF}}+0.5\Delta_{\mathrm{CS}}$. A configuration is considered only if it satisfies
$\Delta_{\mathrm{CF}}\geq0$ and $\Delta_{\mathrm{CS}}\geq-\epsilon_{\mathrm{CS}}$, where $\epsilon_{\mathrm{CS}}$ denotes the maximum permitted CS accuracy decrease.
Tunable baselines use the same development data, selection objective, and
CS-loss constraint. 

\section{Results}

\subsection{RQ1: Directed Prior Attraction}

\begin{figure}[!t]
\centering
\includegraphics[width=0.8\columnwidth]{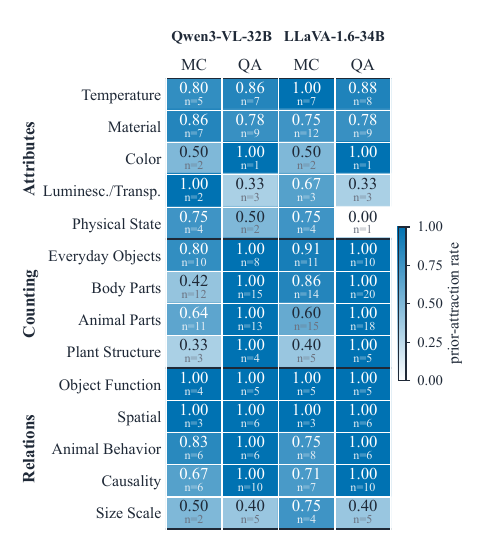}
\caption{\textbf{Prior attraction by CDH subcategory.} Each cell reports the
fraction of original CF errors that select the no-image prior winner. Rows
represent CDH subcategories, and columns represent model--task settings.}
\label{fig:rq1-matrix}
\end{figure}

Figure~\ref{fig:rq1-matrix} reports the prior-attraction rate among incorrect CF predictions across subcategories, models, and tasks, with $n$ denoting the number of CF errors in each cell. Across most cells, incorrect CF predictions tend to select the candidate preferred without the image. This pattern is particularly pronounced for counting and relation categories. Notably, in most cases, the text-only preferred candidate matches the ground-truth answer for the paired CS image, indicating that the errors are directed toward the answer associated with the ordinary CS state rather than an arbitrary frequent answer.

\subsection{RQ2: Comparison with Existing Methods}

Table~\ref{tab:main} reports the performance of SPC on the MC test set. SPC raises CF accuracy by 7.8 and 7.0 points for Qwen and LLaVA, respectively, while changing CS accuracy by only +0.4 and -0.4 points. Repairs substantially outnumber harms (21/3 and 17/1). Consistent with these gains, CFAD and RPD decrease, indicating a narrower CF--CS performance gap, while the lower CCR shows that fewer remaining CF errors converge on the paired CS answer. SPC therefore repairs atypical cases while largely preserving correct predictions on matched CS images.

\begin{table*}[!t]
\centering
\small
\setlength{\tabcolsep}{3.2pt}
\begin{tabular*}{\textwidth}{@{\extracolsep{\fill}}llrrrrrrr@{}}
\toprule
Model & Method & CF-Acc & CS-Acc & CFAD$\downarrow$ & RPD$\downarrow$
& CCR$\downarrow$ & $\Delta_{\mathrm{CF}}$/$\Delta_{\mathrm{CS}} (\%)$ & \#Repairs/harms \\
\midrule
Qwen3-VL-32B
& Original model & 0.665 & 0.965 & 0.300 & 0.311 & 0.857 & -- & -- \\
& \textbf{SPC} & \textbf{0.743} & \textbf{0.970} & \textbf{0.226}
& \textbf{0.233} & \textbf{0.695} & \textbf{+7.8/+0.4} & 21/3 \\
\midrule
LLaVA-1.6-34B
& Original model & 0.565 & 0.917 & 0.352 & 0.384 & 0.830 & -- & -- \\
& \textbf{SPC} & \textbf{0.635} & \textbf{0.913} & \textbf{0.278}
& \textbf{0.305} & \textbf{0.750} & \textbf{+7.0/-0.4} & 17/1 \\
\bottomrule
\end{tabular*}
\vspace{-2mm}
\caption{\textbf{Performance of SPC on MC data with $\epsilon_{\mathrm{CS}}=0.04$.} ``\#Repairs/harms'' gives the numbers of
corrected errors and new errors. The 95\% intervals for $\Delta_{\mathrm{CF}}$/$\Delta_{\mathrm{CS}}$ are
[3.9, 12.2]/[-2.2, 3.0] for Qwen and [3.5, 10.4]/[-2.6, 1.7] for LLaVA.}
\vspace{-4mm}
\label{tab:main}
\end{table*}

Table~\ref{tab:controlled} compares SPC with existing mitigation methods using $U_{0.5}=\Delta_{\mathrm{CF}}+0.5\Delta_{\mathrm{CS}}$ to summarize the repair--retention trade-off. For fairness, the same development objective is used to select every tunable method. The baselines vary substantially across models and tasks: some provide little CF repair, while others reduce CF accuracy or incur notable CS loss. NoLan is the strongest existing baseline, but SPC achieves the highest utility in all four model--task settings. The $\epsilon_{\mathrm{CS}}=0.04$ setting performs best on Qwen MC, while $\epsilon_{\mathrm{CS}}=0.10$ performs best in the other three settings, demonstrating a more consistent balance between CF repair and CS retention. Paired significance tests are reported in the appendix.

\begin{table}[!tb]
\centering
\scriptsize
\setlength{\tabcolsep}{2.0pt}
\resizebox{\columnwidth}{!}{%
\begin{tabular}{lrrrr}
\toprule
& \multicolumn{2}{c}{Qwen3-VL-32B}
& \multicolumn{2}{c}{LLaVA-1.6-34B} \\
\cmidrule(lr){2-3}\cmidrule(lr){4-5}
Method & MC & QA & MC & QA \\
\midrule
Focus on Vision & -0.9/+0.4 & +8.7/-0.4 & -12.6/+3.0 & -13.0/+2.2 \\
VCD & -0.9/+0.0 & +1.3/+0.0 & +2.6/-1.7 & -0.4/+0.0 \\
MFCD & -2.2/+0.4 & +1.7/+0.0 & +3.5/-2.2 & -3.0/-0.4 \\
PAI & +0.4/-0.4 & +3.0/+0.0 & +1.3/+0.0 & +3.9/-0.4 \\
REVIS & -1.3/+0.0 & -2.2/-0.4 & -1.3/+0.9 & +14.8/-3.0 \\
NoLan & +4.8/-1.7 & +15.2/-0.9 & +11.7/-3.5 & +24.8/-8.7 \\
\midrule
SPC ($\epsilon_{\mathrm{CS}}=0.04$, ours) & \textbf{+7.8/+0.4} & +22.2/-1.7
& +7.0/-0.4 & +23.9/-4.8 \\
SPC ($\epsilon_{\mathrm{CS}}=0.10$, ours) & +8.3/-0.9 & \textbf{+23.9/-4.3}
& \textbf{+12.6/-2.2} & \textbf{+30.0/-9.1} \\
\bottomrule
\end{tabular}
}
\vspace{-2mm}
\caption{\textbf{Comparison of CF repair and CS retention across mitigation methods.} Each cell is $\Delta_{\mathrm{CF}}$/$\Delta_{\mathrm{CS}}$ in percentage points. Bold marks the highest $U_{0.5}=\Delta_{\mathrm{CF}}+0.5\Delta_{\mathrm{CS}}$ in each model--task column.}
\vspace{-2mm}
\label{tab:controlled}
\end{table}

\subsection{RQ3: Component Analysis of SPC}

Table~\ref{tab:components} compares SPC with three variants: \emph{No prior subtraction} directly selects the highest-scoring image-conditioned candidate; \emph{Fixed correction strength} uses $\lambda=0.8$ for every instance and directly outputs the highest-scoring corrected candidate; and \emph{Ungated adaptive subtraction} uses the learned instance-dependent correction strength but directly adopts every proposal.

\begin{table}[!tb]
\centering
\scriptsize
\setlength{\tabcolsep}{1.8pt}
\resizebox{\columnwidth}{!}{%
\begin{tabular}{lrrrr}
\toprule
& \multicolumn{2}{c}{Qwen3-VL-32B}
& \multicolumn{2}{c}{LLaVA-1.6-34B} \\
\cmidrule(lr){2-3}\cmidrule(lr){4-5}
Variant & MC & QA & MC & QA \\
\midrule
No prior subtraction ($\lambda=0$) & -0.9/-1.3 & -22.6/-0.4
& -1.3/+1.7 & -3.5/+0.9 \\
Fixed correction strength ($\lambda=0.8$) & +10.4/-2.2 & +17.4/-0.9
& +11.3/-0.9 & +33.0/-13.9 \\
Ungated adaptive subtraction & +9.6/-2.6 & +22.6/-4.8
& +10.9/-0.9 & +30.9/-10.4 \\
Full SPC ($\epsilon_{\mathrm{CS}}=0.04$) & +7.8/+0.4 & +22.2/-1.7
& +7.0/-0.4 & +23.9/-4.8 \\
\bottomrule
\end{tabular}
}
\vspace{-2mm}
\caption{\textbf{Component analysis of SPC.} Each cell is $\Delta_{\mathrm{CF}}$/$\Delta_{\mathrm{CS}}$ in percentage points.}
\label{tab:components}
\vspace{-2mm}
\end{table}

\paragraph{No prior subtraction.}
The original prediction is produced through standard decoding and parsed into candidate set
$\mathcal{Y}$, whereas this variant ranks all candidates using their
image-conditioned teacher-forced scores. The appendix details these
procedures and explains why they may yield different predictions. As shown in
Table~\ref{tab:components}, candidate rescoring without prior subtraction
reduces CF accuracy in all four model--task settings, most substantially on
Qwen QA. Thus, the gains of SPC do not come simply from changing
free-form generation into candidate comparison: prior subtraction provides the corrective effect.

\paragraph{Fixed correction strength.}
As shown in Table~\ref{tab:components}, a fixed coefficient can be competitive, confirming that candidate-specific prior subtraction is itself an effective correction direction. However, its performance is less consistent across model--task settings, and excessive subtraction can remove useful prior support from matched CS instances.

The appendix provides additional comparisons under more challenging settings, including reduced development sets and CDH-category holdouts.  Adaptive correction achieves higher average utility than the corresponding fixed-strength variant across both models in these settings (Appendix Table~\ref{tab:matched-shift}). It also yields positive average CF gains at every development-set size (Appendix Table~\ref{tab:calibration-size}). These results show that instance-dependent correction is particularly valuable when fitting data are limited or the test distribution differs from the fitting distribution.

\paragraph{Ungated adaptive subtraction.}
This variant tests whether adapting the correction strength is sufficient
without selective answer revision. It adopts the highest-scoring corrected
candidate for every instance. Compared with full SPC, it generally achieves
more CF repair but incurs larger losses on matched CS images, particularly in
QA. Therefore, the two components serve complementary roles: the learned
correction strength determines how strongly to adjust the candidate scores, whereas
selective answer revision prevents weak proposals from replacing correct
predictions.

\begin{figure}[!t]
\centering
\includegraphics[width=0.8\columnwidth]{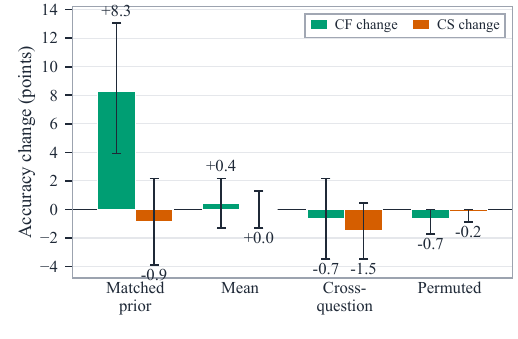}
\caption{\textbf{Effect of prior-direction controls on Qwen3-VL-32B under MC.} Box plots show the distributions over 10,000 pair-level bootstrap resamples.}
\label{fig:evidence-chain}
\end{figure}

We conduct additional ablations to test three design choices: matching each
prior score to its corresponding candidate, fitting with both CF and CS
instances, and estimating prior preference by removing the image. Replacing
the current candidate-specific prior scores with the mean prior over the development set, prior scores borrowed from another question, or a random permutation of the current prior scores across candidates removes nearly all MC repair, as shown in Figure~\ref{fig:evidence-chain}. Fitting
with CF instances alone increases repair but causes substantial loss on
matched CS images, confirming the importance of paired CF--CS fitting.
Replacing the image-removed condition with white or black images yields
similar prior rankings and retains positive gains, showing that the results
do not depend on a single prior-estimation procedure. Full results are
provided in the appendix.

\subsection{RQ4: Generalization of SPC}

We evaluate whether SPC generalizes to held-out CDH categories, unseen candidate-answer orders, and external benchmarks. For CDH category generalization, we exclude one CDH category during fitting and evaluate on the held-out category. For candidate-answer order generalization, we fit using the original candidate order and evaluate on candidate-answer permutations. For generalization to external benchmarks, we select all configurations on CDH-Bench and apply them to other benchmarks without retuning.

\paragraph{Generalization to held-out CDH categories.}
Table~\ref{tab:family-holdout} reports the category-holdout results. SPC
achieves positive CF gains in every model--task setting, showing that the
correction transfers to held-out CDH categories. CS performance remains stable
on MC, whereas QA retention varies more across held-out categories. This greater variation motivates testing SPC's sensitivity to QA question forms
and claim orders in RQ5.

\begin{table}[t]
\centering
\scriptsize
\setlength{\tabcolsep}{2.0pt}
\resizebox{\columnwidth}{!}{%
\begin{tabular}{lrrrr}
\toprule
& \multicolumn{2}{c}{Qwen3-VL-32B}
& \multicolumn{2}{c}{LLaVA-1.6-34B} \\
\cmidrule(lr){2-3}\cmidrule(lr){4-5}
Held-out CDH Category & MC & QA & MC & QA \\
\midrule
Attribute & +9.3/+0.0 & +10.7/-9.3 & +5.3/+0.0 & +5.3/-1.3 \\
Counting & +10.0/+0.0 & +33.8/-3.8 & +10.0/-1.2 & +41.2/-5.0 \\
Relation & +6.7/+4.0 & +29.3/+0.0 & +10.7/-1.3 & +36.0/-12.0 \\
\bottomrule
\end{tabular}
}
\vspace{-2mm}
\caption{\textbf{Generalization to held-out CDH categories.} Each cell reports
$\Delta_{CF}/\Delta_{CS}$ in percentage points.}
\vspace{-5mm}
\label{tab:family-holdout}
\end{table}

\paragraph{Generalization to alternative candidate orders.}
Across three unseen orderings of the same MC candidates, Qwen3-VL-32B and LLaVA-1.6-34B retain
positive CF gains with similar absolute accuracy. Thus, the gains do not
depend on a particular A--D ordering. Detailed results are provided in the
appendix.

\paragraph{Generalization to external benchmarks.}

Table~\ref{tab:spectrum} evaluates whether SPC configured and fitted on CDH-Bench generalizes to external conflict and object-hallucination benchmarks without retuning. CDH-Bench results and NoLan are included as references.  SPC achieves positive gains across the external conflict benchmarks.  Notably, SPC changes almost no predictions on POPE or POPEv2, whereas NoLan changes substantially more.  These results indicate that SPC generalizes selectively when visual evidence conflicts with an identifiable ordinary-state answer, rather than to object-hallucination benchmarks in general. Detailed results are provided in the appendix.

\begin{table*}[!t]
\centering
\scriptsize
\setlength{\tabcolsep}{1.2pt}
\begin{tabular*}{\textwidth}{@{\extracolsep{\fill}}llr rr c rr c rr c@{}}
\toprule
 & & & \multicolumn{3}{c}{SPC ($\epsilon_{\mathrm{CS}}=0.04$, ours)}
& \multicolumn{3}{c}{SPC ($\epsilon_{\mathrm{CS}}=0.10$, ours)}
& \multicolumn{3}{c}{NoLan} \\
\cmidrule(lr){4-6} \cmidrule(lr){7-9} \cmidrule(lr){10-12}
Benchmark & Reported change & \#Instances
& \#Changed & \#Repairs/harms & Effect 
& \#Changed & \#Repairs/harms & Effect 
& \#Changed & \#Repairs/harms & Effect \\
\midrule
\multicolumn{12}{l}{\itshape Conflict benchmarks} \\
CDH-Bench MC & $\Delta_{\mathrm{CF}}$/$\Delta_{\mathrm{CS}}$ & 460
& 40 & 26/7 & \textbf{+7.8/+0.4}
& 49 & 29/12 & \textbf{+8.3/-0.9}
& 53 & 24/17 & +4.8/-1.7 \\
CDH-Bench QA & $\Delta_{\mathrm{CF}}$/$\Delta_{\mathrm{CS}}$ & 460
& 59 & 53/6 & \textbf{+22.2/-1.7}
& 69 & 57/12 & \textbf{+23.9/-4.3}
& 42 & 37/4 & \textbf{+15.2/-0.9} \\
Visual CounterFact & $\Delta_{\mathrm{CF}}$/$\Delta_{\mathrm{CS}}$ & 2440
& 45 & 40/5 & \textbf{+3.1/-0.2}
& 56 & 44/12 & \textbf{+3.4/-0.8}
& 46 & 33/11 & +2.5/-0.7 \\
HallusionBench & $\Delta_{\mathrm{CF}}$/$\Delta_{\mathrm{CS}}$ & 656
& 34 & 24/10 & \textbf{+5.2/-0.9}
& 45 & 28/17 & \textbf{+6.1/-2.7}
& 109 & 36/56 & -2.4/-3.7 \\
ConflictVIS MC & $\Delta_{\mathrm{accuracy}}$ & 374
& 8 & 8/0 & \textbf{+2.1}
& 8 & 8/0 & \textbf{+2.1}
& 8 & 8/0 & \textbf{+2.1} \\
ConflictVIS QA & $\Delta_{\mathrm{accuracy}}$ & 374
& 96 & 96/0 & +25.7
& 96 & 96/0 & +25.7
& 97 & 97/0 & \textbf{+25.9} \\
\midrule
\multicolumn{12}{l}{\itshape Object-hallucination benchmarks} \\
 POPE & $\Delta_{\mathrm{accuracy}}$/$\Delta_{\mathrm{F1}}$ & 9000
& 17 & 6/11 & -0.1/+0.0
& 18 & 6/12 & -0.1/+0.0
& 572 & 299/273 & \textbf{+0.3/+1.0} \\
 POPEv2 & $\Delta_{\mathrm{absent}}$/$\Delta_{\mathrm{present}}$ & 140
& 1 & 0/1 & \textbf{-1.4/+0.0}
& 1 & 0/1 & \textbf{-1.4/+0.0}
& 12 & 0/12 & -17.1/+0.0 \\
\bottomrule
\end{tabular*}
\caption{\textbf{Performance of SPC and NoLan on CDH-Bench and external conflict and object-hallucination benchmarks.} Within each benchmark, bold marks nondominated results among the compared methods.}
\vspace{-5mm}
\label{tab:spectrum}
\end{table*}

\subsection{RQ5: Scope of Prior-Based Correction}

Using the binary QA task, Table~\ref{tab:equivariance} examines the scope of prior-based correction by varying the question form and claim order and measuring how the estimated prior's preference for the ordinary answer relates to CF repair.

\begin{table}[!t]
\centering
\footnotesize
\setlength{\tabcolsep}{3.0pt}
\begin{tabular}{lrrr}
\toprule
Question form & \shortstack{Prior favors \\ ordinary (\%)} &
\shortstack{$\Delta_{CF}$ when prior \\ favors ordinary} &
\shortstack{$\Delta_{CF}$\\otherwise} \\
\midrule
\multicolumn{4}{l}{\itshape Qwen3-VL-32B} \\
\cmidrule(lr){1-4}
Contrastive & 57.4 & +46.2/+22.7 & +22.4/-3.1 \\
Instead & 46.1 & +25.5/+21.7 & +9.7/+0.0 \\
Choice & 49.1 & +8.0/+23.0 & -1.7/-0.9 \\
Reject second & 24.3 & +55.4/+26.8 & +44.3/-6.3 \\
\midrule
\multicolumn{4}{l}{\itshape LLaVA-1.6-34B} \\
\cmidrule(lr){1-4}
Contrastive & 46.5 & +52.3/+14.0 & +41.5/-14.6 \\
Instead & 14.3 & +60.6/+9.1 & +56.3/-14.7 \\
Choice & 5.7 & +23.1/+7.7 & +21.2/-15.2 \\
Reject second & 0.0 & --/-- & +82.2/-43.0 \\
\bottomrule
\end{tabular}
\caption{\textbf{Effect of question form and claim order on prior preference and CF repair.} The percentage column reports instances for which the estimated prior favors the ordinary answer under both claim orders. The last two columns report $\Delta_{\mathrm{CF}}$ when this condition holds and otherwise, with results under the ordinary-first and atypical-first claim orders separated by a slash.}
\vspace{-5mm}
\label{tab:equivariance}
\end{table}

\paragraph{Effect of question form.}
We express the same question in four forms: \emph{Contrastive} asks whether the image supports ``five rather than six'', \emph{Instead} uses ``five instead of six'', \emph{Choice} asks which of the two answers is supported by the image, and \emph{Reject second} asks whether choosing ``six'' is incorrect because ``five'' is correct. Only the Contrastive form is used during fitting; the other three forms are unseen. Table~\ref{tab:equivariance} shows that the estimated prior favors the ordinary answer less frequently for some unseen forms, particularly for LLaVA. Thus, question forms affect how reliably the no-image scores estimate the model's preference for the ordinary answer.

\paragraph{Effect of claim order.}
Each question form is evaluated under two claim orders: ordinary-first places
the ordinary claim first, whereas atypical-first places the atypical claim
first. We classify an instance as having an ordinary-state prior preference
only when the no-image condition predicts ``yes'' in the ordinary-first order
and ``no'' in the atypical-first order. This criterion distinguishes a
preference for the ordinary claim from a fixed preference for either ``yes''
or ``no''. Table~\ref{tab:equivariance} reports the percentage of instances
satisfying this criterion and the corresponding $\Delta_{CF}$ within
and outside this group. When the estimated prior consistently favors the
ordinary state, SPC improves CF accuracy under both claim orders. Outside this
group, the gains are concentrated in the ordinary-first order and generally
disappear or become negative when the claims are reversed, particularly for
LLaVA. Thus, prior-based correction is reliable only when
the estimated prior remains associated with the ordinary state. Additional
statistics are provided in the appendix.

\section{Discussion}

\paragraph{Relation to NoLan.}
NoLan and SPC both contrast model scores obtained with and without the image,
but differ in two respects. First, NoLan modifies token probabilities during
generation using a hand-designed rule, whereas SPC adjusts candidate-answer
scores over a fixed candidate set using a learned instance-dependent correction
strength. Second, SPC retains the original prediction when the corrected answer
lacks sufficient support, while NoLan applies its correction directly during
decoding. Empirically, NoLan achieves more CF repair in some settings but shows
less consistent CS retention and changes substantially more predictions on
POPE and POPEv2. Overall, SPC provides a more conservative correction that
explicitly balances CF repair against CS retention.

\paragraph{Limitations.}
SPC requires a fixed candidate set, access to candidate-level probabilities,
and model-specific paired CF--CS development data. During inference, it scores
each candidate both with and without the image, making it more expensive than
one-pass decoding and incompatible with interfaces that expose only generated
text. The current method therefore does not support open-ended generation.
Table~\ref{tab:equivariance} further reveals a limitation on binary QA: the
no-image preference may reflect a fixed yes/no bias or claim position rather
than a preference for the ordinary state. Although the final decision rule can
reject weak corrections, it cannot compensate for a misaligned prior estimate.
Future work will extend SPC to open-ended generation, reduce its scoring cost,
and improve the stability of prior estimation across question forms and claim
orders.

\section{Conclusion}

This work identifies directed prior attraction as a systematic error pattern in
commonsense-driven hallucination (CDH) and introduces SPC to mitigate it. Experiments show that SPC effectively reduces CDH while largely preserving correct predictions on ordinary cases.
Further analysis demonstrates its generalization and clarifies that reliable
prior estimation is essential for successful correction.

\clearpage
\onecolumn
\appendix
\setcounter{figure}{0}
\setcounter{table}{0}
\setcounter{equation}{0}
\renewcommand{\thefigure}{A\arabic{figure}}
\renewcommand{\thetable}{A\arabic{table}}
\renewcommand{\theequation}{A\arabic{equation}}
\setcounter{tocdepth}{2}
\makeatletter
\def\addcontentsline#1#2#3{%
  \addtocontents{#1}{\protect\contentsline{#2}{#3}{\thepage}{}%
    \protected@file@percent}}
\makeatother

\begin{center}
{\LARGE\bfseries Supplementary Material}\\[0.75em]
{\Large When Model Priors Conflict with Visual Evidence:}\\[0.15em]
{\Large Mitigating Commonsense-Driven Hallucinations by Selective Prior
Calibration}\\[0.75em]
{\large Kesheng Chen, Yamin Hu, Wenjian Luo}
\end{center}

\renewcommand{\contentsname}{Contents}
\tableofcontents
\clearpage
\twocolumn

\section{Benchmark and Evaluation}

\subsection{Benchmark Composition and Data Split}

CDH-Bench contains 300 matched counterfactual (CF) and commonsense (CS) pairs
in 14 subcategories
\cite{chen2026cdh}. We reserve five pairs from each subcategory for fitting and
model selection. This gives 70 development pairs and 230 test pairs. This
section describes the data and the information available to every method.

We use the same terms as the main paper. The \emph{prior-favored candidate} is
the candidate ranked highest by the text-only scores. \emph{Directed prior
attraction} is the tendency of incorrect CF responses to move toward this
candidate. In CDH, the prior-favored candidate often expresses the ordinary
state.

\begin{figure*}[t]
\centering
\setlength{\tabcolsep}{2pt}
\textbf{Counting (100 pairs)}\\
\begin{tabular}{@{}cccc@{}}
\begin{tabular}{@{}cc@{}}
\includegraphics[width=0.102\textwidth,height=0.068\textwidth,keepaspectratio]{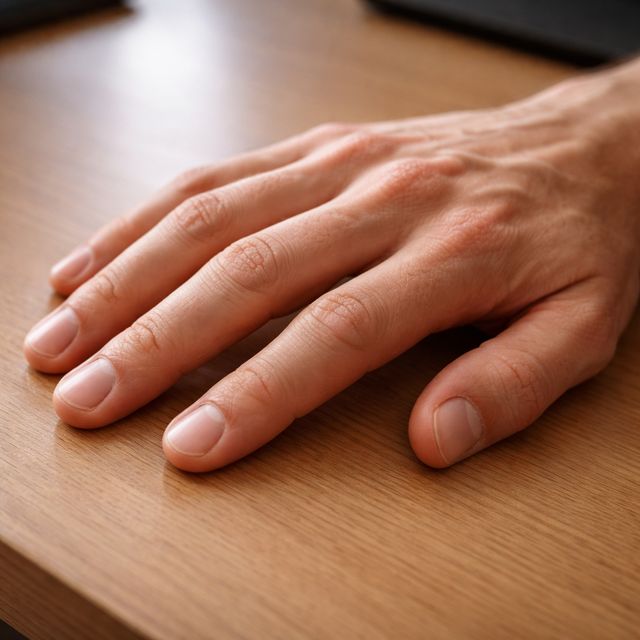} &
\includegraphics[width=0.102\textwidth,height=0.068\textwidth,keepaspectratio]{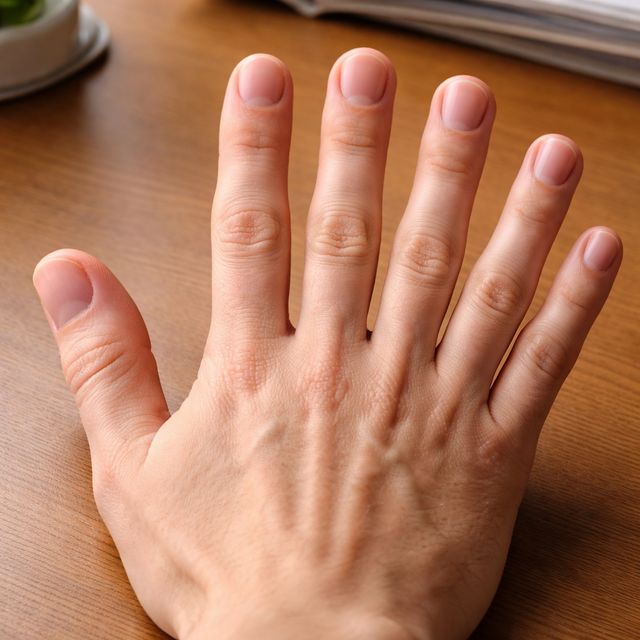} \\
\multicolumn{2}{c}{\small Body Parts}
\end{tabular} &
\begin{tabular}{@{}cc@{}}
\includegraphics[width=0.102\textwidth,height=0.068\textwidth,keepaspectratio]{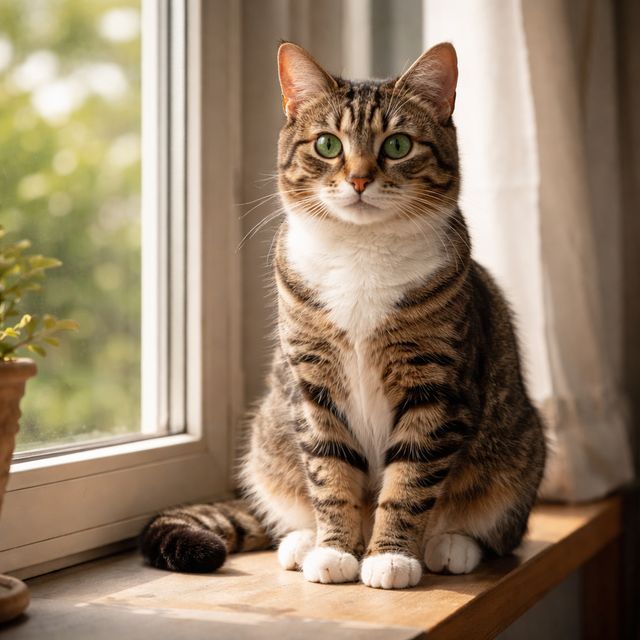} &
\includegraphics[width=0.102\textwidth,height=0.068\textwidth,keepaspectratio]{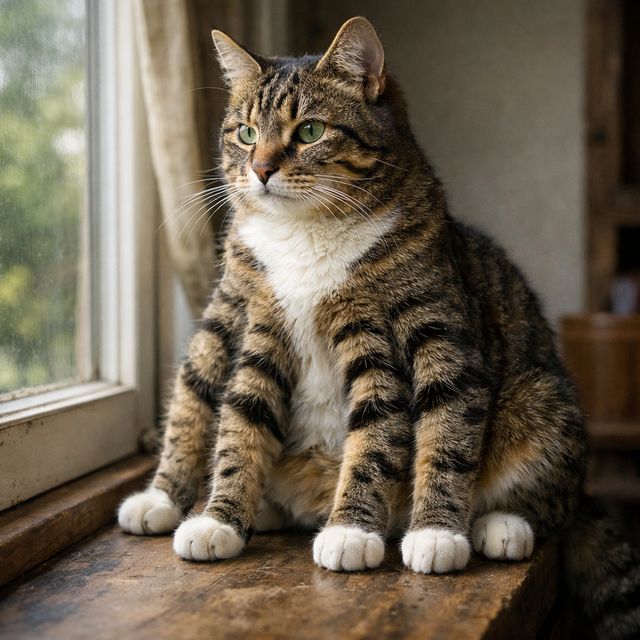} \\
\multicolumn{2}{c}{\small Animal Parts}
\end{tabular} &
\begin{tabular}{@{}cc@{}}
\includegraphics[width=0.102\textwidth,height=0.068\textwidth,keepaspectratio]{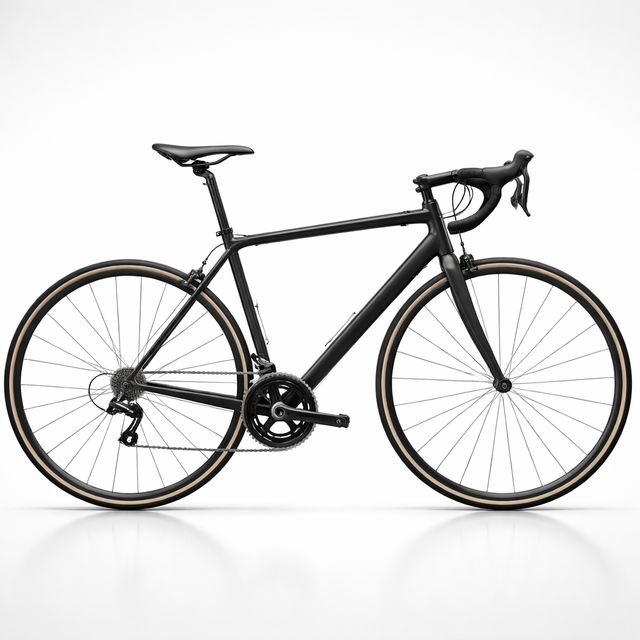} &
\includegraphics[width=0.102\textwidth,height=0.068\textwidth,keepaspectratio]{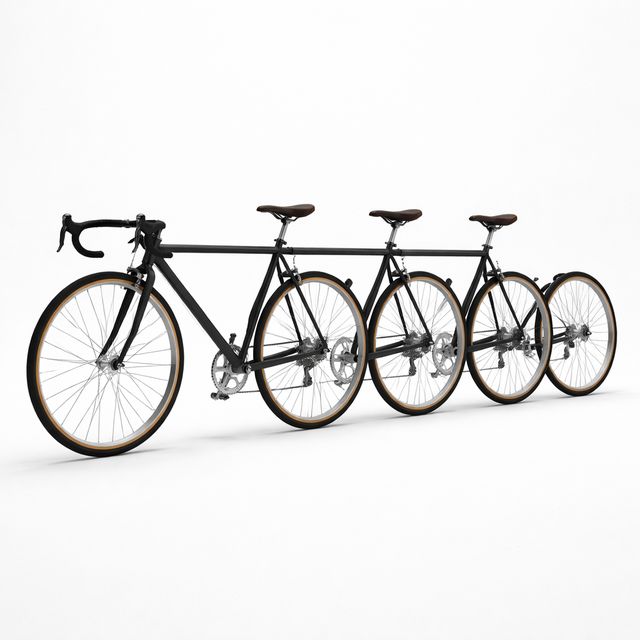} \\
\multicolumn{2}{c}{\small Everyday Objects}
\end{tabular} &
\begin{tabular}{@{}cc@{}}
\includegraphics[width=0.102\textwidth,height=0.068\textwidth,keepaspectratio]{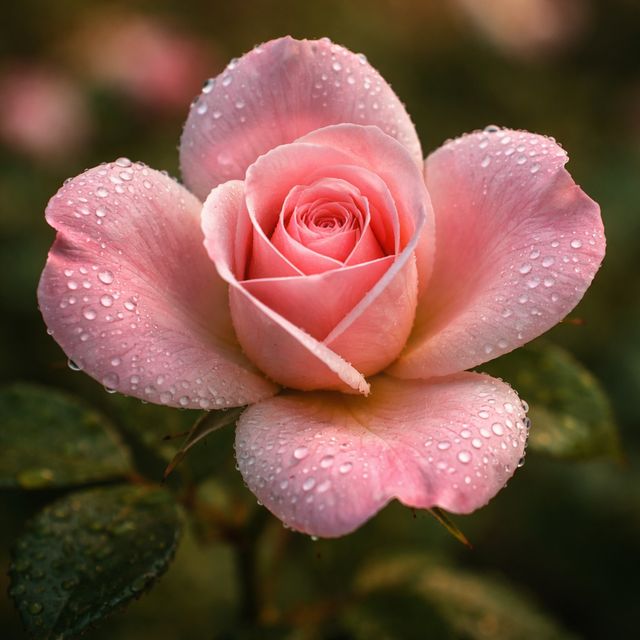} &
\includegraphics[width=0.102\textwidth,height=0.068\textwidth,keepaspectratio]{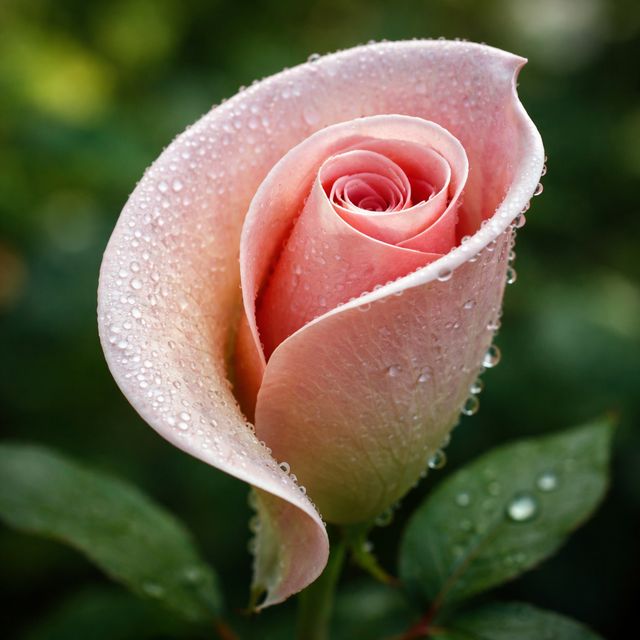} \\
\multicolumn{2}{c}{\small Plant Structure}
\end{tabular}
\end{tabular}\\[3pt]
\textbf{Attributes (100 pairs)}\\
\begin{tabular}{@{}ccccc@{}}
\begin{tabular}{@{}cc@{}}
\includegraphics[width=0.081\textwidth,height=0.060\textwidth,keepaspectratio]{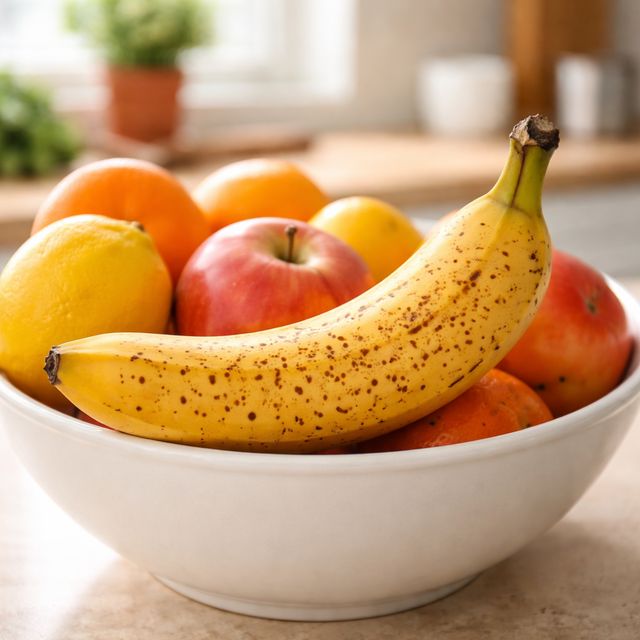} &
\includegraphics[width=0.081\textwidth,height=0.060\textwidth,keepaspectratio]{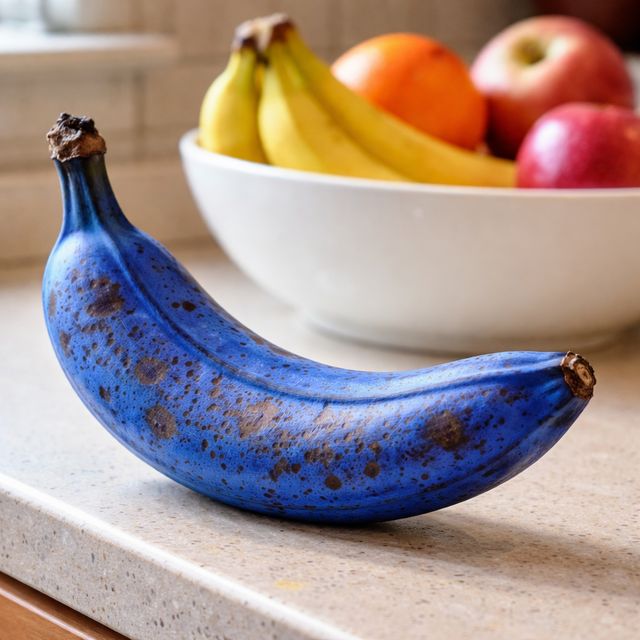} \\
\multicolumn{2}{c}{\small Color}
\end{tabular} &
\begin{tabular}{@{}cc@{}}
\includegraphics[width=0.081\textwidth,height=0.060\textwidth,keepaspectratio]{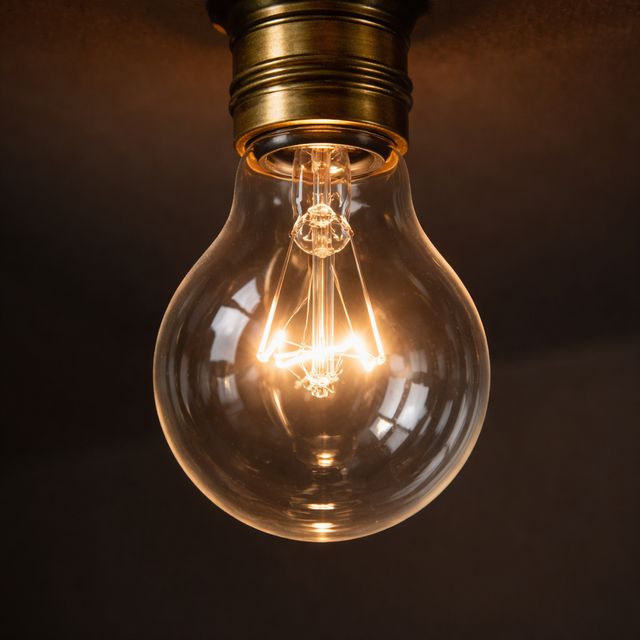} &
\includegraphics[width=0.081\textwidth,height=0.060\textwidth,keepaspectratio]{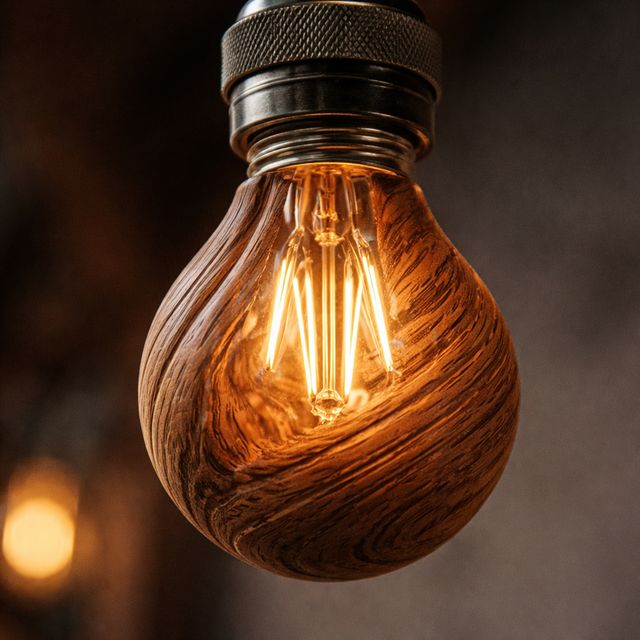} \\
\multicolumn{2}{c}{\small Material}
\end{tabular} &
\begin{tabular}{@{}cc@{}}
\includegraphics[width=0.081\textwidth,height=0.060\textwidth,keepaspectratio]{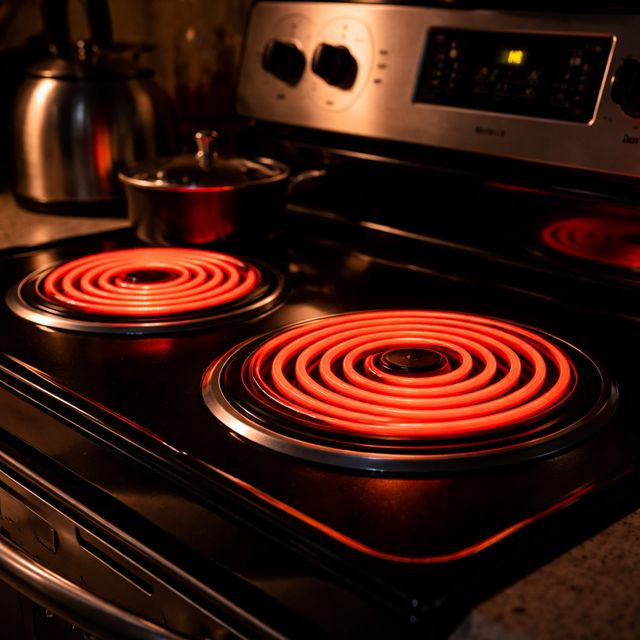} &
\includegraphics[width=0.081\textwidth,height=0.060\textwidth,keepaspectratio]{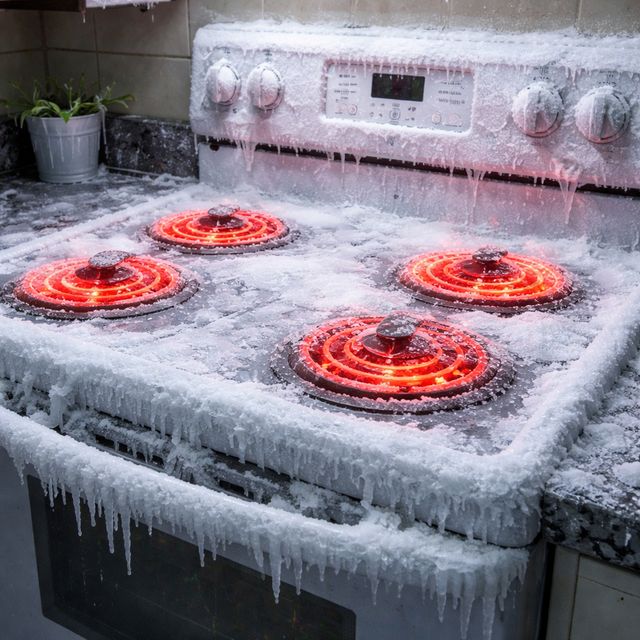} \\
\multicolumn{2}{c}{\small Temperature}
\end{tabular} &
\begin{tabular}{@{}cc@{}}
\includegraphics[width=0.081\textwidth,height=0.060\textwidth,keepaspectratio]{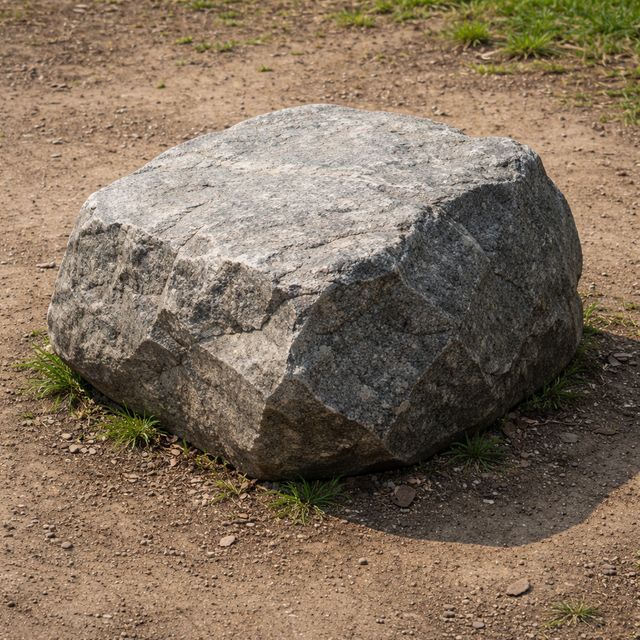} &
\includegraphics[width=0.081\textwidth,height=0.060\textwidth,keepaspectratio]{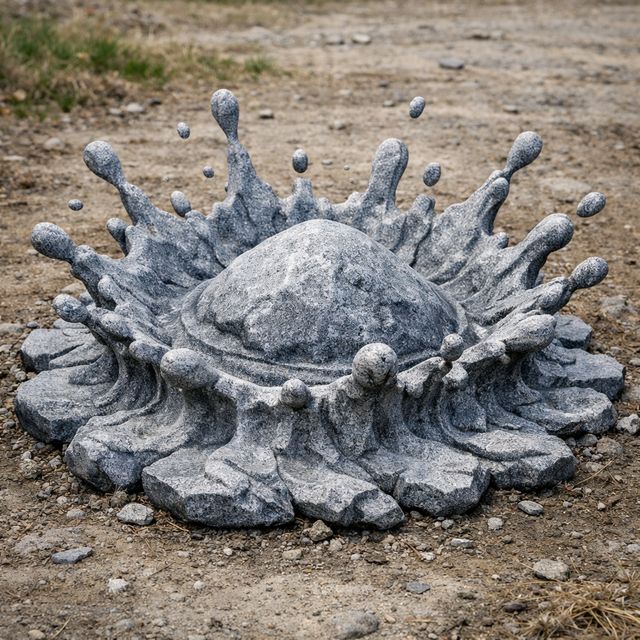} \\
\multicolumn{2}{c}{\small Physical State}
\end{tabular} &
\begin{tabular}{@{}cc@{}}
\includegraphics[width=0.081\textwidth,height=0.060\textwidth,keepaspectratio]{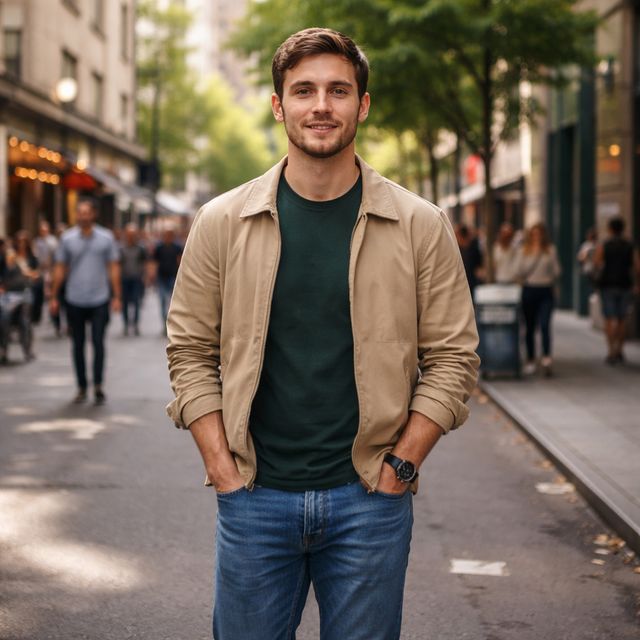} &
\includegraphics[width=0.081\textwidth,height=0.060\textwidth,keepaspectratio]{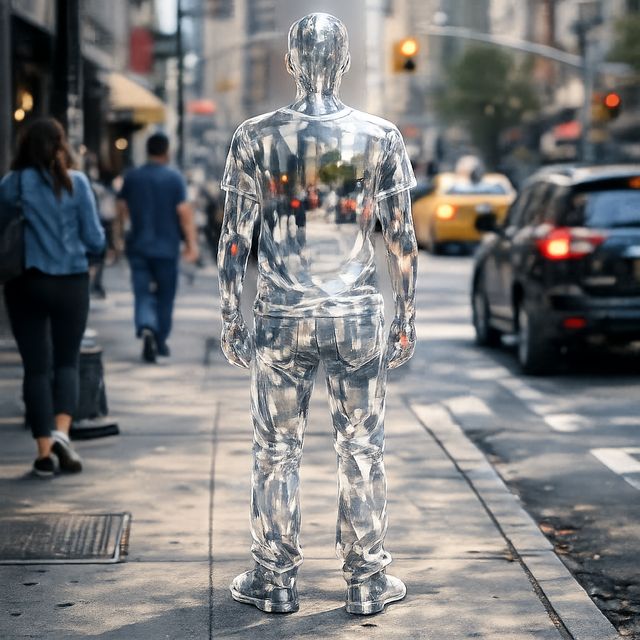} \\
\multicolumn{2}{c}{\small Luminescence/Transparency}
\end{tabular}
\end{tabular}\\[3pt]
\textbf{Relations (100 pairs)}\\
\begin{tabular}{@{}ccccc@{}}
\begin{tabular}{@{}cc@{}}
\includegraphics[width=0.081\textwidth,height=0.060\textwidth,keepaspectratio]{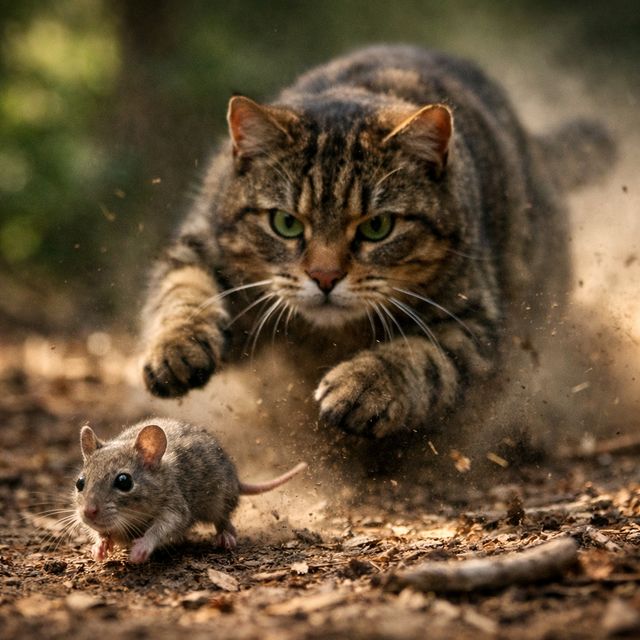} &
\includegraphics[width=0.081\textwidth,height=0.060\textwidth,keepaspectratio]{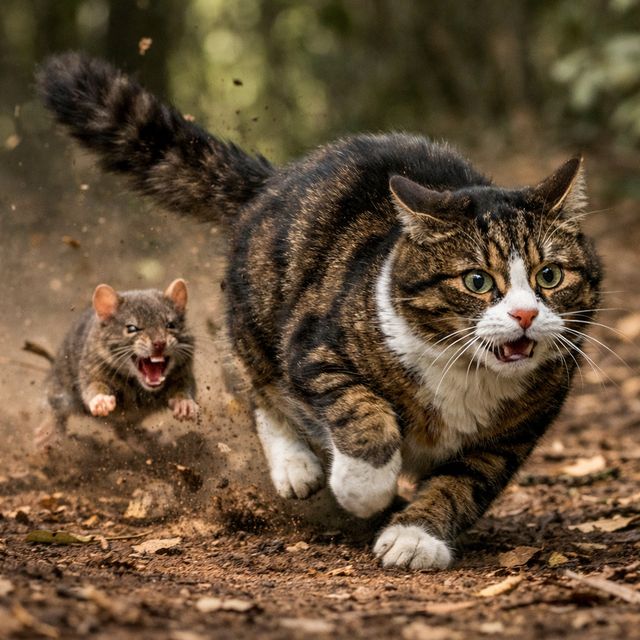} \\
\multicolumn{2}{c}{\small Animal Behavior}
\end{tabular} &
\begin{tabular}{@{}cc@{}}
\includegraphics[width=0.081\textwidth,height=0.060\textwidth,keepaspectratio]{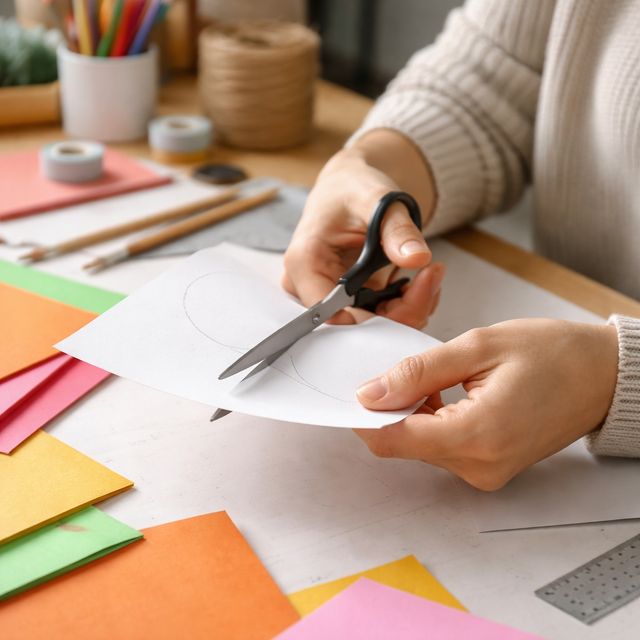} &
\includegraphics[width=0.081\textwidth,height=0.060\textwidth,keepaspectratio]{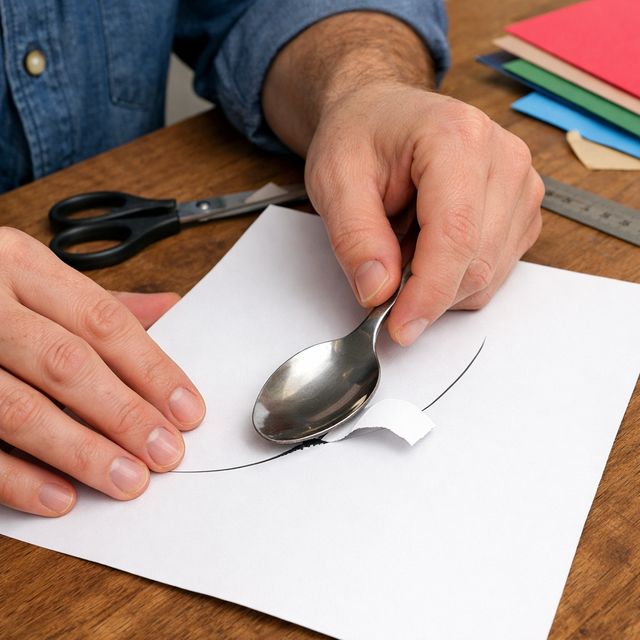} \\
\multicolumn{2}{c}{\small Object Function}
\end{tabular} &
\begin{tabular}{@{}cc@{}}
\includegraphics[width=0.081\textwidth,height=0.060\textwidth,keepaspectratio]{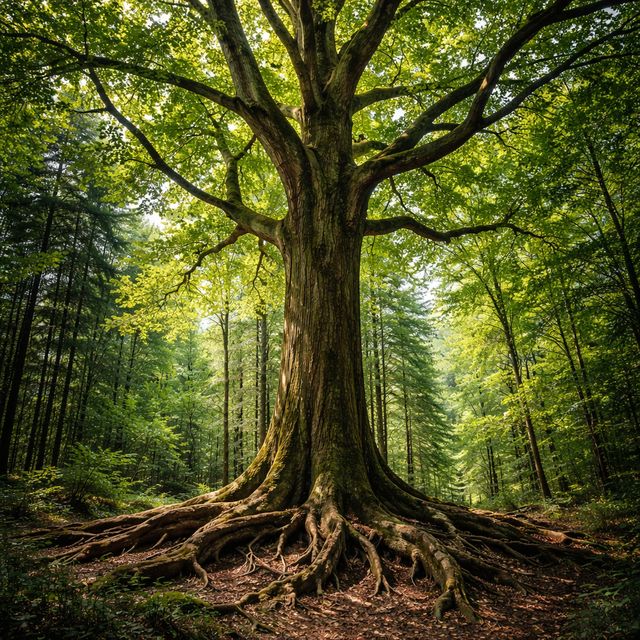} &
\includegraphics[width=0.081\textwidth,height=0.060\textwidth,keepaspectratio]{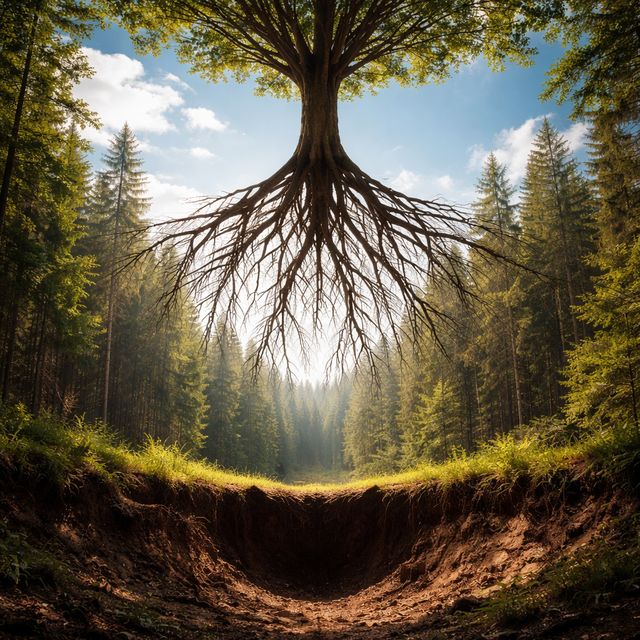} \\
\multicolumn{2}{c}{\small Spatial}
\end{tabular} &
\begin{tabular}{@{}cc@{}}
\includegraphics[width=0.081\textwidth,height=0.060\textwidth,keepaspectratio]{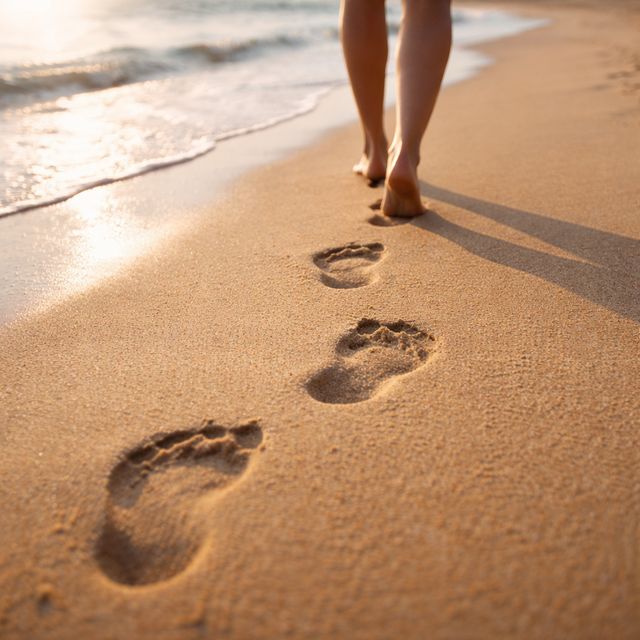} &
\includegraphics[width=0.081\textwidth,height=0.060\textwidth,keepaspectratio]{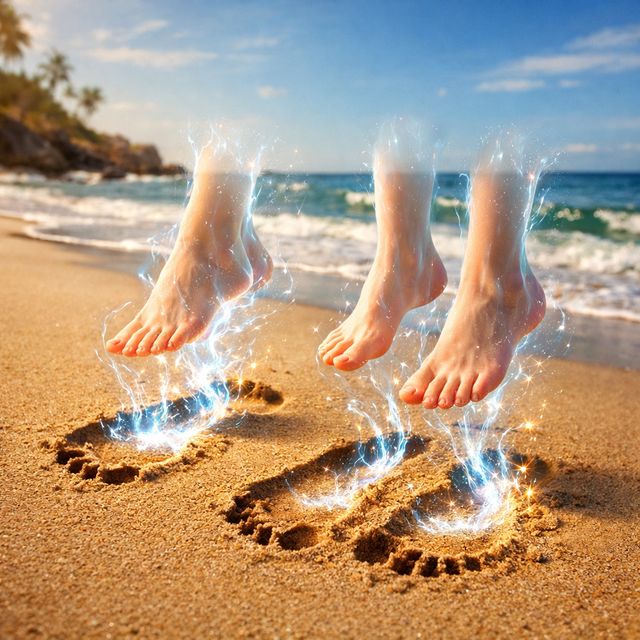} \\
\multicolumn{2}{c}{\small Causality}
\end{tabular} &
\begin{tabular}{@{}cc@{}}
\includegraphics[width=0.081\textwidth,height=0.060\textwidth,keepaspectratio]{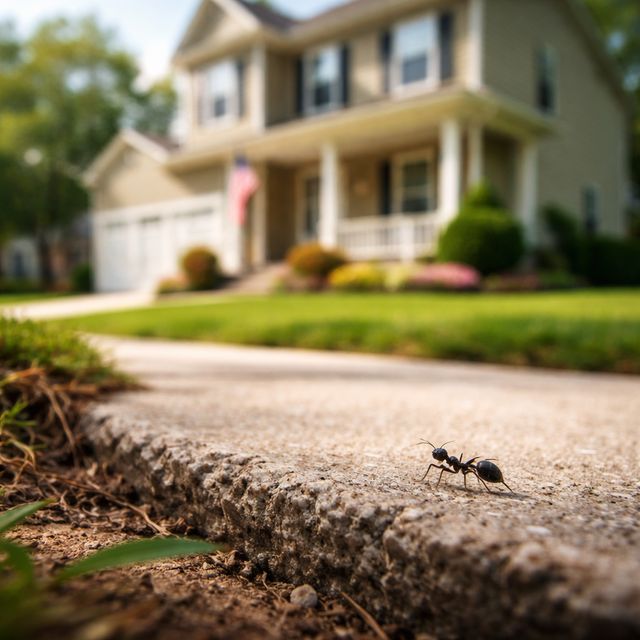} &
\includegraphics[width=0.081\textwidth,height=0.060\textwidth,keepaspectratio]{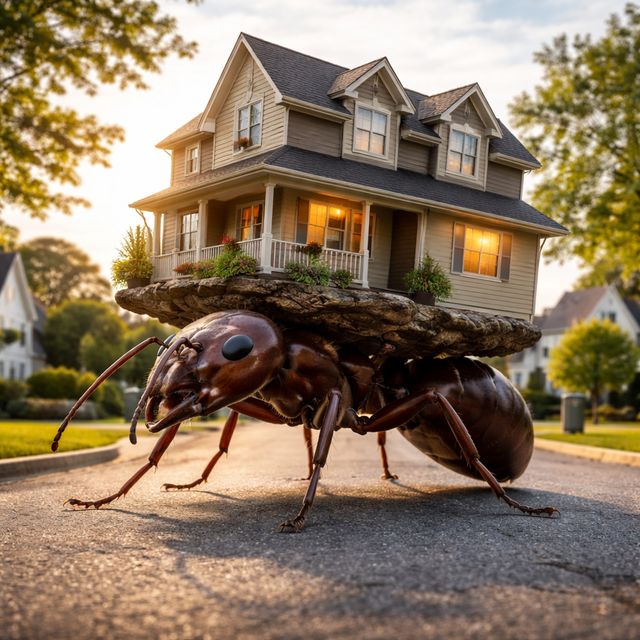} \\
\multicolumn{2}{c}{\small Size Scale}
\end{tabular}
\end{tabular}
\caption{\textbf{Representative CDH pairs from all 14 subcategories.} Within every
cell, the matched CS image is on the left and the CF image is on the right. The
three categories change what must be recovered from the image: counts,
visual attributes, or the relation between entities and events. Every test
prediction still receives only one of the two images.}
\label{fig:supp-category-gallery}
\end{figure*}

Figure~\ref{fig:supp-category-gallery} shows examples beyond counting.
Counting contains four subcategories with 25 pairs each. Attributes contain
Color and Material with 25 pairs each, Temperature with 20 pairs, and Physical
State and Luminescence/Transparency with 15 pairs each. Relations contain five
subcategories with 20 pairs each. Each pair keeps the queried concept fixed.
It changes whether the image agrees with the ordinary state. The trade-off
between CF repair and CS retention therefore covers counts, attributes, and
relations.

The 70 development pairs are chosen by a fixed rule. We take the first five
numbered pairs in each of the 14 subcategories. The remaining 230 pairs form
the test set. Both images and both answer formats from a pair always stay in
the same split. The split is fixed before fitting or selecting any setting.

\subsection{Tasks and Evaluation Settings}

Multiple choice (MC) is the primary test of answer-level correction because candidate positions
can change without changing their meaning. Binary QA serves a different role.
Its original wording strongly favors ``no'' on CF images and ``yes'' on CS
images. We therefore use QA to test whether a gain survives changes in claim
order and question form.

As in the main paper, $\epsilon_{\mathrm{CS}}$ denotes the largest allowed
CS accuracy decrease on development data. Most results use
$\epsilon_{\mathrm{CS}}=0.04$. We also report
$\epsilon_{\mathrm{CS}}=0.10$, which allows a larger CS decrease. A second
version accounts for uncertainty in the learned correction strength.
Qwen3-VL-32B \cite{bai2025qwen3vl} and LLaVA-1.6-34B
\cite{liu2024llavabaselines} are evaluated on MC, QA, the balanced claim-order
test, and external transfer benchmarks.
We also evaluate Qwen3-VL-30B-A3B \cite{bai2025qwen3vl} on MC. For this model,
we test error direction, paired repair and retention, and transfer to CDH
categories excluded from fitting. We use the same 70/230 split.

RQ5 introduces a balanced claim-order test. It asks the same visual question
with the ordinary and atypical answers in both orders. This test separates a
preference for the ordinary answer from a fixed preference for ``yes'' or
``no.'' Section~\ref{sec:supp-rq5} gives the full construction and question
forms.

\subsection{Prompts, Decoding, and Candidate Scoring}

Original MC predictions use the benchmark question followed by its four
options and the instruction ``Answer with a single letter (A, B, C, or D).''
Original QA predictions use the benchmark question followed by ``Answer with
yes or no.'' Decoding is greedy with temperature zero and at most 64 new
tokens. We remove any completed thinking block and parse the first valid
answer: A--D for MC or yes/no for QA. An output with no valid answer is
incorrect.

Candidate scoring keeps the same question, options, and answer instruction. It
then adds ``Final answer:'' and scores the supplied continuations
\texttt{ A}, \texttt{ B}, \texttt{ C}, and \texttt{ D} for MC or
\texttt{ yes} and \texttt{ no} for QA. A candidate score is the sum of its
teacher-forced token log-probabilities. In the text-only condition, we omit the
image content from the same chat template; the text is unchanged. Qwen images
are resized to $512\times512$ pixels. LLaVA uses its model-provided image
processor.

\subsection{Inputs During Online Inference}
Every example is evaluated independently. At inference, SPC receives one
image, its question, and the candidates supplied by MC or QA. It cannot access
the paired image, CF/CS identity, category, pair identifier, or test answer.
Removing the image creates the text-only scoring condition for that example.
A white, black, or mean image gives an alternative prior estimate. None of
these conditions provides a second view of the scene.

\section{Why Correction Strength Depends on the Example}

SPC subtracts a prior-preference estimate for each candidate. The useful
correction strength depends on how strongly the prior affects that decision.
Too little leaves a prior-favored candidate dominant when it conflicts with
the image. Too much can erase useful support on a CS example. The following
analysis makes this trade-off explicit. Let $\mathcal{Y}$ be a fixed set of
possible answers, and let $y^*$ be the answer supported by visual evidence. We
model the centered
image-conditioned score with three terms: visual evidence $v(y)$, prior
contribution $\alpha p(y)$, and residual error $\epsilon(y)$, where
$\alpha\geq0$:
\begin{equation}
\bar s_I(y)=v(y)+\alpha p(y)+\epsilon(y),\qquad
\bar s_P(y)=p(y).
\end{equation}
This is an idealized decomposition used for intuition. In practice,
$\bar s_P$ is a text-only proxy for $p$. Any difference between that proxy and
the latent prior is included in $\epsilon$.
SPC ranks $s_\lambda(y)=\bar s_I(y)-\lambda\bar s_P(y)$. In the ideal case,
$\epsilon(y)=0$ and $\lambda=\alpha$. Candidate-specific subtraction then
removes the prior term exactly. We next ask how far $\lambda$ can differ from
$\alpha$ while preserving the visually supported answer.

\paragraph{When subtraction recovers the visual answer.}
Define
\begin{align}
\Delta_v &= \min_{y\ne y^*}\bigl[v(y^*)-v(y)\bigr],\\
R_p &= \max_{y\ne y^*}|p(y^*)-p(y)|,\\
R_e &= \max_{y\ne y^*}|\epsilon(y^*)-\epsilon(y)|.
\end{align}
If $\Delta_v>R_e$, $R_p>0$, and
\begin{equation}
|\lambda-\alpha|<(\Delta_v-R_e)/R_p,
\end{equation}
then $y^*$ is the unique top candidate under SPC.

For every $y\ne y^*$, the corrected pairwise margin satisfies
\begin{align}
s_\lambda(y^*)-s_\lambda(y)
&=v(y^*)-v(y)\\
&\quad +(\alpha-\lambda)[p(y^*)-p(y)]\\
&\quad +\epsilon(y^*)-\epsilon(y)\\
&\ge \Delta_v-|\lambda-\alpha|R_p-R_e>0.
\end{align}
When $R_p=0$, prior subtraction cannot alter any pairwise ranking and the
condition reduces to $\Delta_v>R_e$.

\paragraph{Why one correction strength cannot work for every example.}
For sample $j$ with $\Delta_{v,j}>R_{e,j}$ and $R_{p,j}>0$, let
$\delta_j=(\Delta_{v,j}-R_{e,j})/R_{p,j}$ and
$A_j=(\alpha_j-\delta_j,\alpha_j+\delta_j)\cap[0,\infty)$. A single fixed
correction strength meets the condition for every sample only when
$\bigcap_j A_j$ is not empty. If these ranges do not overlap, no fixed strength
can satisfy this sufficient condition for every sample. A fixed strength may
still work outside the bound because the bound is conservative. An
instance-dependent model can place each
$\lambda_j$ inside its own $A_j$ when the observed scores distinguish the
samples. This motivates learning the correction strength instead of assigning
one value to every question.

The quantities $\Delta_v$, $R_p$, and $R_e$ explain the decision, but the model
does not expose them separately. SPC does not attempt to recover their true
values. Image-conditioned and text-only margins instead describe how sharply
the candidates are separated. Entropy describes uncertainty. Differences
between the distributions and their top answers indicate how strongly the
estimated prior can alter the ranking. The learned feature map uses these
observed relations to predict a
correction strength. The range analysis explains why this strength should
depend on the observed scores. It does not assume that the three score terms
can be separated exactly.

The same idea can be checked directly from the observed scores. Each condition
$s_\lambda(y)\ge s_\lambda(y')$ is linear in $\lambda$. Together, these
conditions define the exact range over which $y$ is ranked first. SPC uses the
strength predicted by its learned correction model. We also evaluate a version
that accounts for uncertainty in correction strength. It measures the probability
assigned to this range instead of treating the predicted strength as exact.

\begin{figure}[t]
\centering
\includegraphics[width=0.9\columnwidth]{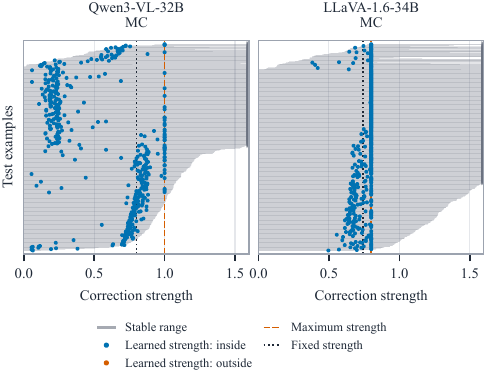}
\caption{\textbf{Stability of corrected proposals as correction strength
changes.}
Each line is one MC test example. Segments show the range over which the
proposal remains unchanged. Examples are sorted by the lower and then upper
endpoint of this range. Ranges are shown only up to 1.6. Dots show the
fitted $\lambda_\theta(u)$. The dashed line shows the selected maximum
correction strength. The dotted line shows the constant $\lambda^*$ selected
on development data.}
\label{fig:lambda-intervals}
\end{figure}

\paragraph{How stable are the corrected proposals?}
Figure~\ref{fig:lambda-intervals} shows how the corrected proposal changes
with the correction strength. The median width of the range that keeps the
same proposal is
1.10--1.74 across the four model and task combinations. For 15--51\% of
examples, the range has no upper bound. The selected maximum correction
strength stays within the range for
76.5\% of Qwen MC examples and 99.6\% of LLaVA MC examples. Whenever
$\lambda_\theta$ reaches this maximum, it remains inside the stable range.

We also replace $\lambda_\theta(u)$ with a constant $\lambda^*$ selected on the
development set. The constants are 0.80 for Qwen and 0.74 for LLaVA. They fall
outside the stable ranges on 5.9\% and 1.1\% of examples. Range width does not
distinguish successful repairs from harms. Define the stability radius as
$r=\min(\lambda-\lambda_{\mathrm{lo}},\lambda_{\mathrm{hi}}-\lambda)$.
Repaired and harmed examples have nearly identical median $r$ values, 0.29 and
0.28. Thus, the radius alone does not separate useful changes from harmful
ones. Small changes in correction strength leave most proposals unchanged.
Selective answer revision still needs information beyond the width of the
stable range.

Selective answer revision cannot remove every trade-off. Suppose two examples
have identical score features, but the same answer change repairs one and harms
the other. A fixed rule must treat them alike. Paired development data
therefore choose a statistical balance between CF repair and CS retention.

These results explain the two decisions in SPC. The learned strength handles
variation in prior strength. Selective answer revision handles ambiguity after
correction. Paired CF and CS outcomes fit the correction model and select the
revision threshold.

\section{How SPC Is Fitted and Applied}

SPC learns how strongly to subtract candidate-level prior preference. It
selects when to adopt the resulting proposal on development data. This section
gives the complete feature map, fitting objective, setting selection, and
online inference procedure.

\subsection{Learning the Correction Strength}

The evaluated implementation uses a regularized linear model. Its inputs
describe the candidate score distributions for one example. Fitting uses paired
CF and CS examples from a development set $D$, which show both successful repairs and harmful
corrections. At inference, the model uses image-conditioned scores, text-only
scores, the original answer, and the candidate count.

Let the observable input to the correction model be
$o=(\bar s_I,\bar s_P,i_b,K)$. Let $K=|\mathcal Y|$,
$q_I=\operatorname{softmax}(\bar s_I)$,
$q_P=\operatorname{softmax}(\bar s_P)$, and let $m_I,m_P$ be the margins
between the two highest scores. Write $i_I,i_P,i_b$ for the positions of the
answers ranked first by
the image-conditioned scores, the text-only scores, and the original prediction,
respectively. The feature vector is
\begin{align}
u=\phi(o)=[&1,\log K/\log4,\mathsf H(q_I)/\log K,\mathsf H(q_P)/\log K,\nonumber\\
&\max q_I,\max q_P,\tanh(m_I/4),\tanh(m_P/4),\nonumber\\
&JS(q_I,q_P),\mathbf1[i_I=i_P],\nonumber\\
&\mathbf1[i_b=i_I],\mathbf1[i_b=i_P],
\tanh((m_P-m_I)/4)].
\end{align}
Here $\mathsf H$ denotes entropy. $JS$ denotes Jensen--Shannon divergence,
which measures the difference between two probability distributions.
The first two values are a bias term and the normalized candidate count.
The next six describe entropy, peak probability, and the margin between the
two highest scores under
the two scoring conditions. The remaining values describe distribution
disagreement, agreement among the answers ranked first, and the margin gap
$\tanh((m_P-m_I)/4)$. Together, the features summarize candidate count,
confidence, and image--prior disagreement. They form the evaluated linear
correction model. The control using
image-conditioned scores alone removes every input that depends on the prior
estimate. We standardize each nonconstant feature using its mean and standard
deviation on $D$; a zero standard deviation is replaced by one. We leave the
bias feature unchanged. In the equations below, $u$ denotes this standardized
feature vector. The fitting and inference strengths are
\begin{equation}
\begin{aligned}
\lambda_\theta^{\mathrm{fit}}(u)
 &=\operatorname{softplus}(u^\top\theta),\\
\lambda_\theta(u)
 &=\min\{\lambda_\theta^{\mathrm{fit}}(u),\lambda_{\max}\}.
\end{aligned}
\end{equation}

For a development set $D$, SPC fits a regularized correction model with
$\theta\sim\mathcal N(0,\rho^{-1}I)$ and minimizes
\begin{equation}
\sum_{i\in D}-\log\operatorname{softmax}
  (\bar s_{I,i}-\lambda_\theta^{\mathrm{fit}}(u_i)\bar s_{P,i})_{y_i}
  +\frac{\rho}{2}\|\theta\|_2^2.
\end{equation}
The regularization strength $\rho$ is selected on the development set. The
maximum-a-posteriori (MAP) estimate $\hat\theta$ defines the standard SPC
correction model.

\subsection{Uncertainty-Aware Revision under Distribution Shift}

The category-holdout, candidate-order, and reduced-data experiments change the
conditions seen during fitting. For these tests, we use an uncertainty-aware
extension of SPC. It averages corrected probabilities over plausible
correction strengths. It also checks whether the proposal stays first across
those strengths and whether the score pattern resembles the development data.
The fitting objective and candidate scores remain unchanged.

Let
$J(\theta)$ denote the objective above and
$H=\nabla_\theta^2J(\hat\theta)$. We form a Laplace approximation
$\theta\sim\mathcal N(\hat\theta,\Sigma)$. Here, $\Sigma$ is the symmetrized
pseudoinverse of $H+10^{-8}I$. Its eigenvalues are clipped to
$[10^{-8},4]$ for numerical stability. For a test vector $u$,
\begin{equation}
\eta=u^\top\theta\sim\mathcal N(\mu,\sigma^2),\quad
\mu=u^\top\hat\theta,\quad \sigma^2=u^\top\Sigma u.
\end{equation}
We set $\lambda(\eta)=\min\{\operatorname{softplus}(\eta),
\lambda_{\max}\}$ and compute
\begin{equation}
\widetilde q(y)=
\mathbb E_{\eta}\!\left[
\operatorname{softmax}\bigl(\bar s_I-\lambda(\eta)\bar s_P\bigr)_y
\right]
\label{eq:supp-posterior-predictive}
\end{equation}
with 64-point Gauss--Hermite quadrature. The proposal is
$z_U=\arg\max_y\widetilde q(y)$, and its margin is
$m_U=\widetilde q(z_U)-\max_{y\ne z_U}\widetilde q(y)$.

We also compute the exact range of strengths for which $z_U$ stays first:
\begin{equation}
\begin{aligned}
\mathcal L(z_U)
={}&\{\lambda\in[0,\lambda_{\max}]:\\
&\bar s_I(z_U)-\lambda\bar s_P(z_U)
\geq \bar s_I(y)-\lambda\bar s_P(y),\ \forall y\}.
\end{aligned}
\end{equation}
Its posterior support is
$r_U=\Pr[\lambda(\eta)\in\mathcal L(z_U)]$. Each pairwise inequality is
linear in $\lambda$, so $\mathcal L(z_U)$ is obtained by intersecting its
closed half-intervals.

The extension can also reject score patterns far from the development data.
Let $x$ contain the 12 standardized nonbias features, let $C_D$ be their
development-set covariance, and define
\begin{equation}
d_{\mathrm{dev}}=\sqrt{x^\top(C_D+0.1I)^{+}x}.
\end{equation}
We retain the two support conditions from the main paper, replacing the point
proposal $z_M$ with $z_U$. Thus,
$d=(i_I\ne i_P)\land(z_U=i_I)$ and
$a=(i_b=i_I=i_P)\land(z_U\ne i_b)$. The complete decision is
\begin{align}
g_U={}&(z_U\ne i_b)\land(K\in\mathcal K_D)\nonumber\\
&{}\land(m_U\geq m_{\min})\land(d_{\mathrm{dev}}\leq d_{\max})\nonumber\\
&{}\land\bigl[(d\land r_U\geq\tau_D)
\lor(a\land r_U\geq\tau_A)\bigr].
\label{eq:supp-uncertainty-gate}
\end{align}
Here, $\tau_D$ and $\tau_A$ are the minimum posterior support for conditions
$d$ and $a$. The threshold $m_{\min}$ is the minimum proposal margin, and
$d_{\max}$ is the largest accepted development distance. The set
$\mathcal K_D$ contains the candidate counts observed in $D$.
The final answer is $z_U$ when $g_U$ is true and $i_b$ otherwise.
Standard SPC uses the point estimate $\hat\theta$, no distance check, and the
decision in the main paper. Results use standard SPC unless a caption
explicitly names the uncertainty-aware extension.

\subsection{Selecting Settings on Development Data}

For each model, $D$ contains MC and QA data from the same 70 pairs. The CF and
CS sides in both tasks contribute four examples per pair. Model selection
chooses the highest mean value of
$\Delta_{\mathrm{CF}}+0.5\Delta_{\mathrm{CS}}$ across MC and QA. Each task
must have a CF change of at least zero and satisfy
$\Delta_{\mathrm{CS}}\geq-\epsilon_{\mathrm{CS}}$. The fitted model is
model-specific but receives no task indicator.
Table~\ref{tab:spc-primary-settings} lists the settings selected for the
primary results.

\begin{table}[t]
\centering
\small
\setlength{\tabcolsep}{2.0pt}
\begin{tabular*}{\columnwidth}{@{\extracolsep{\fill}}p{0.58\columnwidth}cc@{}}
\toprule
Setting & \shortstack{Qwen3-VL-\\32B} & \shortstack{LLaVA-1.6-\\34B} \\
\midrule
Regularization $\rho$ & 0.01 & 0.01 \\
Maximum correction strength & 1.0 & 0.8 \\
Minimum gap between the top two corrected-answer probabilities & 0.3 & 0.2 \\
Candidate counts accepted at test time & 2, 4 & 2, 4 \\
\bottomrule
\end{tabular*}
\caption{\textbf{Settings selected for each model with
$\epsilon_{\mathrm{CS}}=0.04$.} These are the standard point-estimate SPC
settings used for the primary results.}
\label{tab:spc-primary-settings}
\end{table}

We select $\rho$ from $\{0.01,0.1,1,10\}$. We select the maximum correction
strength from $\{0.8,1,1.2,1.5\}$. Both choices use
$U_{0.5}=\Delta_{\mathrm{CF}}+0.5\Delta_{\mathrm{CS}}$. The minimum proposal
margin is selected from $\{0,0.05,0.1,0.2,0.3\}$.

For standard SPC with $\epsilon_{\mathrm{CS}}=0.10$, Qwen selects
$\rho=0.01$, $\lambda_{\max}=1.0$, and $m_{\min}=0$. LLaVA selects
$\rho=1.0$, $\lambda_{\max}=1.0$, and $m_{\min}=0$. The
uncertainty-aware extension searches $\tau_D$ and $\tau_A$ over
$\{0.5,0.65,0.8,0.9,0.95,1.01\}$. It searches $d_{\max}$ over the 90th, 95th,
and 99th percentiles of the development distances. For compactness, define
$h=(\rho,\lambda_{\max},\tau_D,\tau_A,m_{\min},d_{\max})$. The selected
values are:
\begin{center}
\scriptsize
\setlength{\tabcolsep}{2.5pt}
\begin{tabular}{@{}lll@{}}
\toprule
Model & $\epsilon_{\mathrm{CS}}$ & $h$ \\
\midrule
Qwen & 0.00, 0.02 & $(0.01,0.8,0.5,1.01,0,6.60)$ \\
 & 0.04 & $(0.01,1.0,0.5,0.5,0.3,6.60)$ \\
 & 0.06, 0.10 & $(0.01,1.0,0.5,0.5,0,6.60)$ \\
LLaVA & 0.00 & $(0.01,0.8,0.5,1.01,0,3.56)$ \\
 & 0.02 & $(0.01,0.8,0.5,0.5,0.3,3.56)$ \\
 & 0.04 & $(0.01,1.0,0.5,0.8,0.2,6.39)$ \\
 & 0.06 & $(1.0,0.8,0.5,0.5,0.05,6.39)$ \\
 & 0.10 & $(1.0,1.2,0.5,0.5,0,6.39)$ \\
\bottomrule
\end{tabular}
\end{center}
All values are fixed before test evaluation.

\subsection{Applying SPC During Online Inference}
During online inference, SPC first records the original answer. It uses teacher forcing
to score every supplied answer under the image-conditioned and text-only
conditions. It then computes $u=\phi(o)$ and predicts
$\lambda_\theta(u)$. Next, it subtracts the estimated prior preference for
each answer and forms a corrected proposal. Selective answer revision adopts
the proposal only when it satisfies the conditions in the main paper. No
parameter fitting or threshold selection occurs on test examples.

The uncertainty-aware extension replaces the point proposal and final decision
with Eqs.~\ref{eq:supp-posterior-predictive} and
\ref{eq:supp-uncertainty-gate}. Otherwise, it follows the same procedure.

\section{RQ1: Directed Prior Attraction}

The subcategory matrix in the main paper shows that incorrect CF predictions
often select the prior-favored candidate. The overall rates show the same
pattern in every model and task combination. For Qwen, the rates are 70.1\% on MC
and 90.4\% on QA. For LLaVA, they are 77.0\% and 91.6\%. CDH errors therefore
follow the same direction across both answer formats and models.

The number of errors in each matrix cell matters when reading these rates.
Several attribute cells and a few MC counting cells contain very few original
CF errors. Their rates are less stable. Outside cells with few errors, the
attraction is clear and is strongest for counting and relation QA. In most
cells, the prior-favored candidate is also the correct answer for the matched
CS image. The direction is therefore tied to the ordinary answer rather than
to an arbitrary frequent label.

\section{RQ2: Comparison with Existing Methods}

Table~\ref{tab:official-cdh-full} reports all CDH metrics for the primary
setting. On MC, the paired bootstrap 95\% intervals for
$\Delta_{\mathrm{CF}}/\Delta_{\mathrm{CS}}$ are
[3.9, 12.2]/[-2.2, 3.0] for Qwen and
[3.5, 10.4]/[-2.6, 1.7] for LLaVA. Thus, both models show a clear CF gain and
a CS change near zero. On MC, CFAD, RPD, and CCR decrease for both models.
On QA, CFAD and RPD decrease, while CCR remains at 1.0.

\begin{table}[t]
\centering
\scriptsize
\setlength{\tabcolsep}{1.5pt}
\resizebox{\columnwidth}{!}{%
\begin{tabular}{lllrrrrr}
\toprule
Model & Task & Method & CF Acc. & CS Acc. & CFAD$\downarrow$
& RPD$\downarrow$ & CCR$\downarrow$ \\
\midrule
Qwen3-VL-32B & MC & Original & 0.665 & 0.965 & 0.300 & 0.311 & 0.857 \\
 & & SPC & 0.743 & 0.970 & 0.226 & 0.233 & 0.695 \\
 & QA & Original & 0.591 & 0.948 & 0.357 & 0.376 & 1.000 \\
 & & SPC & 0.813 & 0.930 & 0.117 & 0.126 & 1.000 \\
\midrule
LLaVA-1.6-34B & MC & Original & 0.565 & 0.917 & 0.352 & 0.384 & 0.830 \\
 & & SPC & 0.635 & 0.913 & 0.278 & 0.305 & 0.750 \\
 & QA & Original & 0.535 & 0.922 & 0.387 & 0.420 & 1.000 \\
 & & SPC & 0.774 & 0.874 & 0.100 & 0.114 & 1.000 \\
\midrule
Qwen3-VL-30B-A3B & MC & Original & 0.635 & 0.948 & 0.313 & 0.330 & 0.869 \\
 & & SPC & 0.730 & 0.904 & 0.174 & 0.192 & 0.694 \\
\bottomrule
\end{tabular}
}
\caption{\textbf{Complete CDH-Bench metrics with
$\epsilon_{\mathrm{CS}}=0.04$.} CFAD is
CS Acc. minus CF Acc. RPD normalizes CFAD by CS Acc. CCR is the fraction of
incorrect CF answers that select the paired CS answer. CCR reaches 1.0 on the
original ordinary-first question form. RQ5 therefore evaluates both claim
orders separately.}
\label{tab:official-cdh-full}
\end{table}

\subsection{Compared Methods}

All methods follow the protocol in \emph{Inputs During Online Inference}.
Their adjustable parameters are selected on the common 70-pair development set.

We compare five types of method. Focus on Vision uses a grounding prompt
inspired by vision--knowledge conflict analysis \cite{liu2025insight}. VCD
\cite{leng2024vcd} and MFCD \cite{liu2025mfcd} contrast visual views. PAI
\cite{liu2024pai} modifies visual attention. NoLan suppresses language priors
during decoding \cite{ren2026nolan}. REVIS changes hidden states
\cite{wu2026revis}.
NoLan uses the token-level decoder in Eq.~\ref{eq:supp-token-nolan}. VCD,
MFCD, and PAI retain their released token-level equations and fixed parameters
in implementations for each evaluated model. SPC subtracts the estimated prior
preference for each answer in the current question.

\paragraph{NoLan decoding.}
At generation step $t$, let $\ell^I_t$ and $\ell^P_t$ be the logits over all
tokens from the image-conditioned and text-only decoding runs. NoLan
\cite{ren2026nolan} computes
\begin{align}
\gamma_t&=\tfrac12\{D_{\mathrm{KL}}(q^I_t\Vert q^P_t)
                  +D_{\mathrm{KL}}(q^P_t\Vert q^I_t)\},\nonumber\\
\alpha_t&=\beta[\tanh(1/\gamma_t)+1],\qquad \beta=0.8,\nonumber\\
\ell^{N}_t&=(1+\alpha_t)\ell^I_t-\alpha_t\ell^P_t.
\label{eq:supp-token-nolan}
\end{align}
Each run keeps its own decoder state. Both receive the same selected token, so
they keep the same generated prefix. We update $\gamma_t$ and $\alpha_t$
after every token. Greedy decoding uses $\ell^N_t$. NoLan and the matched
$\beta=0$ reference use the same 64-token limit.

We use the released decoding equations and reported parameters. VCD
uses $\alpha=1.0$, $\beta=0.1$, diffusion step 500, and seed 42. MFCD uses
Gaussian high/low views with thresholds 0.1/0.9, unit contrast weights, and
$\beta=0.3$, without its JSD or entropy extensions.
PAI uses attention weight 0.2, CFG scale 1.1, plausibility threshold 0.1, and
the released relative layer span from decoder layer 2 through the final layer.
NoLan uses $\beta=0.8$. Each decoding baseline applies its low-probability token
filter before greedy token selection. The two decoding runs share the same
generated prefix. Each baseline uses the same prompt and parser as the
original model.

The fixed grounding prompt prepends the following text to the original
benchmark prompt: ``Ground your answer in this single image only. Before
deciding, silently verify the task-focused visible count, identity, attribute,
material, state, relation, cause, or scale. Treat normal commonsense defaults
as hypotheses, not facts. If the image conflicts with what is typical, follow
the image. Do not use paired examples, dataset labels, reference images,
ground-truth claims, or any information outside the task prompt and image.''
It then appends the same task-specific output rule used by the original model.
The separate CFG score control uses the candidate scores again. VCD, MFCD, and
PAI use the token-level decoding rules described above.

Our REVIS implementation follows the released direction extraction, layer
search, risk calibration, and hidden-state correction \cite{wu2026revis}.
Direction extraction uses 100 released examples, and risk calibration uses at
most 300 released examples. The risk scale is 1.0, and the threshold is the
85th percentile of factual calibration risk. The selected Qwen and LLaVA
layers are 46 and 56, with thresholds $-0.2178$ and $-0.1068$. Their
paper-reported correction strengths are 1.6 and 1.1. No CDH test label is used
for extraction, calibration, or layer selection. We also evaluate the released
code default of 2.5.

\subsection{Baseline Performance on CDH-Bench}

The token-level baselines make smaller and less consistent changes than SPC.
On Qwen, VCD changes MC/QA by -0.9/+0.0 and +1.3/+0.0 CF/CS points. MFCD
changes them by -2.2/+0.4 and +1.7/+0.0. PAI changes them by +0.4/-0.4 and
+3.0/+0.0. On LLaVA, the VCD changes are +2.6/-1.7 and -0.4/+0.0. The MFCD
changes are +3.5/-2.2 and -3.0/-0.4. The PAI changes are +1.3/+0.0 and
+3.9/-0.4. These methods sometimes repair CDH. Their effects, however, vary
across models and answer formats.

NoLan \cite{ren2026nolan} uses $\beta=0.8$ without CDH tuning. Every
cross-method table measures its changes from the same original predictions as
the other methods. Under this comparison, it changes Qwen MC by +0.0/-4.8
CF/CS points and QA by +13.9/-0.9. The corresponding LLaVA changes are
+9.6/-2.6 and +24.8/-8.7.

We also compare NoLan with a matched decoding loop that sets $\beta=0$. This
comparison isolates the NoLan token update, although its reference predictions
can differ from the common original decoder. The matched changes are
+4.8/-1.7 and +15.2/-0.9 on Qwen, and +11.7/-3.5 and +24.8/-8.7 on LLaVA.

The grounding prompt changes Qwen MC/QA by -0.9/+0.4 and +8.7/-0.4.
For LLaVA, it changes them by -12.6/+3.0 and -13.0/+2.2.
The separate CFG control applied to scores for complete answers selects
$\gamma=1.1/1.5$ for Qwen MC/QA and $\gamma=2.0$ for both LLaVA tasks.
It changes Qwen MC/QA by +0.9/-0.9 and -17.8/-0.4. The corresponding LLaVA
changes are +6.5/+0.9 and +21.3/-4.3. This control does not include PAI's
attention edit or token-level rule. It tests a fixed contrast between
image-conditioned and text-only answer scores.

Allowing a larger CS decrease gives the second standard SPC setting in the
main paper. With $\epsilon_{\mathrm{CS}}=0.10$, Qwen changes MC/QA by
+8.3/-0.9 and +23.9/-4.3. LLaVA changes them by +12.2/-1.3 and
+30.0/-9.6. The uncertainty-aware extension gives the same Qwen changes and
+12.6/-2.2 and +30.0/-9.1 on LLaVA. The larger CS allowance increases CF
repair in all four model--task combinations. Each development-set CS change
remains within the selected limit.

\subsection{Why REVIS's LLaVA QA Gain Does Not Transfer to MC}

With the selected correction strengths, REVIS changes Qwen by -1.3/+0.0 points on MC
and -2.2/-0.4 on QA. On LLaVA, it changes MC by -1.3/+0.9 and QA by
+14.8/-3.0. Every changed LLaVA QA answer moves from yes to no. This includes
36 CF answers and seven CS answers. MC repair instead requires changes among
different answer options. This directional bias explains why the QA gain does
not transfer to MC. Qwen does not show the same QA gain.

\section{RQ3: Component Analysis of SPC}

RQ3 follows the three parts of SPC in the main paper: candidate-specific prior
subtraction, instance-dependent correction strength, and selective answer
revision. It also tests how prior estimation, paired fitting, and candidate
scoring affect these parts.

\subsection{Why Candidate-Specific Prior Scores Matter}

Teacher-forced image scoring alone ($\lambda=0$) replaces the original
prediction with the top image-conditioned candidate. It changes Qwen by
-0.9/-1.3 points on MC and -22.6/-0.4 on QA. It changes LLaVA by -1.3/+1.7
and -3.5/+0.9. Candidate rescoring alone therefore does not explain SPC's CF
gain.

Each control changes the prior scores while keeping the correction model and
selection procedure unchanged. Table~\ref{tab:supp-attribution} reports the
resulting accuracy changes.
\begin{table}[t]
\centering
\scriptsize
\setlength{\tabcolsep}{1.4pt}
\begin{tabular}{@{}L{0.36\columnwidth}*{4}{C{0.135\columnwidth}}@{}}
\toprule
 & \multicolumn{2}{c}{MC} & \multicolumn{2}{c}{QA} \\
Score used in correction & $\Delta_{\mathrm{CF}}$ & $\Delta_{\mathrm{CS}}$ &
$\Delta_{\mathrm{CF}}$ & $\Delta_{\mathrm{CS}}$ \\
\midrule
No prior subtraction ($\lambda=0$) & -0.9 & -1.3 & -22.6 & -0.4 \\
\shortstack[l]{Use image-conditioned scores\\as the prior estimate} & 0.0 & 0.0 & 0.0 & 0.0 \\
Mean prior scores from development set & +0.4 & 0.0 & +20.9 & -5.7 \\
\shortstack[l]{Prior scores from another\\question} & -0.7$\pm$1.2 & -1.5$\pm$1.1 & +17.4$\pm$1.9 & -4.0$\pm$1.0 \\
\shortstack[l]{Prior scores shuffled\\across answers} & -0.7$\pm$0.8 & -0.2$\pm$0.3 & 0.0$\pm$0.0 & 0.0$\pm$0.0 \\
\textbf{SPC} &
\textbf{+8.3} & \textbf{-0.9} & \textbf{+23.9} & \textbf{-4.3} \\
\bottomrule
\end{tabular}
\caption{\textbf{Effect of replacing the prior scores used by SPC on
Qwen3-VL-32B.} Results use the 230-pair test split. Controls that shuffle scores report mean $\pm$
standard deviation over 20 fixed seeds. Entries report
$\Delta_{\mathrm{CF}}$ or $\Delta_{\mathrm{CS}}$ in percentage points. Every
control uses $\epsilon_{\mathrm{CS}}=0.10$. Bold marks
the highest $U_{0.5}$ in each task.}
\label{tab:supp-attribution}
\end{table}

All Qwen controls fit SPC again with the same search ranges and development data.
One control replaces each example's prior scores with an average for the same
task and candidate count. Another uses scores from a different development
question with the same task and candidate count. A third randomly changes the order of scores across
answers in the current example. The two controls that move scores across examples use 20
seeds. The final control subtracts the image-conditioned score vector. It also
removes all features that depend on the prior estimate.

Candidate-specific prior scores give +8.3/-0.9 points on MC and
+23.9/-4.3 on QA. On MC, they give a larger gain than mean prior scores
with paired $p=0.001$. They also outperform every run that uses another
question, with $p\leq0.003$. Repeating the controls with
$\epsilon_{\mathrm{CS}}=0.04$ gives the same ordering. At
$\epsilon_{\mathrm{CS}}=0.10$, every replacement has at most a 0.4-point mean
MC gain, compared with 8.3 points for SPC. At
$\epsilon_{\mathrm{CS}}=0.04$, every replacement has mean MC repair at or
below zero. This result shows that the useful direction comes
from matching each prior score to the same candidate in the current question.

\subsection{Alternative Prior Estimates}
\label{sec:neutral-prior-controls}

SPC targets the model's language and world knowledge preference over the
supplied candidates. Removing the image is one way to estimate this
preference. We also test white, black, and mean images. The mean image uses the
140 unique development images.

Let $p^{(r)}$ be the centered candidate-score vector from prior estimate $r$,
where $r$ is text-only, white, black, or mean image. We normalize each vector by
its root-mean-square value. The normalized-median estimate takes the median
for each candidate and restores the median scale. The
three-input median applies the same operation to text-only, white, and black
scores. The 25\% adjustment is
\begin{equation}
p_{\mathrm{adj}}=0.75p^{(\mathrm{text})}+0.25p_{\mathrm{med}}.
\end{equation}
The disagreement-weighted version reduces this adjustment for candidates on
which the estimates differ. Let $\widetilde p^{(r)}$ denote a normalized score
and let
\begin{align}
\delta(y)&=\operatorname{median}_{r}
\left|\widetilde p^{(r)}(y)
-\operatorname{median}_{r'}\widetilde p^{(r')}(y)\right|,\\
w(y)&=\frac{0.1}{0.1+\delta(y)}.
\end{align}
It uses
$p^{(\mathrm{text})}(y)+0.25w(y)
[p_{\mathrm{med}}(y)-p^{(\mathrm{text})}(y)]$ and then centers the resulting
vector. Table~\ref{tab:neutral-prior-controls} fixes the
SPC settings selected with $\epsilon_{\mathrm{CS}}=0.04$, changing
only how prior preference is estimated.

\begin{table}[t]
\centering
\footnotesize
\setlength{\tabcolsep}{2.0pt}
\begin{tabular}{@{}L{0.48\columnwidth}C{0.22\columnwidth}C{0.22\columnwidth}@{}}
\toprule
\shortstack[l]{Input used to estimate\\prior preference} & MC $\Delta_{\mathrm{CF}}/\Delta_{\mathrm{CS}}$ & QA $\Delta_{\mathrm{CF}}/\Delta_{\mathrm{CS}}$ \\
\midrule
Text-only (image removed) & +7.8/+0.4 & \textbf{+22.2/-1.7} \\
White image & +6.5/-0.9 & +15.7/-0.4 \\
Black image & +6.1/+0.0 & +14.8/+0.0 \\
\shortstack[l]{Mean image from\\development set} & +3.9/+0.9 & +0.9/+0.0 \\
Median of normalized scores & +6.1/-0.9 & +15.7/+0.0 \\
Median of text-only/white/black & +6.5/-1.3 & +16.1/-0.4 \\
25\% adjustment toward median & \textbf{+7.8/+0.9} & +22.2/-2.6 \\
\shortstack[l]{Adjustment weighted by\\estimate disagreement} & +7.4/+0.4 & +22.2/-2.6 \\
\bottomrule
\end{tabular}
\caption{\textbf{Alternative ways to estimate prior preference with
Qwen3-VL-32B on 230 test pairs.} SPC settings use
$\epsilon_{\mathrm{CS}}=0.04$. Entries report
$\Delta_{\mathrm{CF}}/\Delta_{\mathrm{CS}}$ in percentage points. Bold marks the highest
$U_{0.5}=\Delta_{\mathrm{CF}}+0.5\Delta_{\mathrm{CS}}$ in each task.}
\label{tab:neutral-prior-controls}
\end{table}

The text-only top answer agrees with the top answer from a white image on
87.4\% of examples. It agrees with the top answer from a black image on
88.0\%. White and black
images agree
on 96.3\%. The mean development image agrees with text-only scoring on only
76.3\%. An image formed by averaging pixels is therefore not a reliable average visual
input. Combining the estimates does not improve CF repair. We therefore use
text-only scoring as the default prior estimate.

\subsection{Why Paired CF and CS Fitting Matters}

The main paper separates answer scoring, learned prior subtraction, and
selective answer revision. Table~\ref{tab:paired-calibration-full} tests which
paired development outcomes are needed. All three variants use the same
image-conditioned and text-only scores. They also use the same correction
model and search values. They differ only in the examples used for fitting and
setting selection. Test labels are never used for selection.

\begin{table}[t]
\centering
\footnotesize
\setlength{\tabcolsep}{1.5pt}
\begin{tabular}{@{}L{0.31\columnwidth}C{0.31\columnwidth}C{0.31\columnwidth}@{}}
\toprule
\shortstack[l]{Use of CF and CS\\development data} &
\shortstack{MC $\Delta_{\mathrm{CF}}/\Delta_{\mathrm{CS}}$\\(CF repairs/harms)} &
\shortstack{QA $\Delta_{\mathrm{CF}}/\Delta_{\mathrm{CS}}$\\(CF repairs/harms)} \\
\midrule
\multicolumn{3}{l}{\itshape Qwen3-VL-32B} \\
Fit and select on CF and CS & +7.83/+0.43 (21/3) & +22.17/-1.74 (53/2) \\
\shortstack[l]{Fit on CF and CS;\\select on CF} & +8.26/-0.87 (24/5) & +23.91/-4.35 (57/2) \\
Fit and select on CF & +9.57/-58.70 (34/12) & +34.35/-83.04 (82/3) \\
\midrule
\multicolumn{3}{l}{\itshape LLaVA-1.6-34B} \\
Fit and select on CF and CS & +6.96/-0.43 (17/1) & +23.91/-4.78 (57/2) \\
\shortstack[l]{Fit on CF and CS;\\select on CF} & +12.17/-1.30 (29/1) & +30.00/-9.57 (71/2) \\
Fit and select on CF & +16.96/-38.26 (44/5) & +40.87/-79.57 (97/3) \\
\bottomrule
\end{tabular}
\caption{\textbf{Effect of CF and CS data during fitting and setting
selection.} Results use 230 test pairs. Selection with both sides uses
$\epsilon_{\mathrm{CS}}=0.04$ for each task. Selection with CF alone does not
observe CS correctness. Entries are in percentage points.}
\label{tab:paired-calibration-full}
\end{table}

Fitting only on CF examples makes the method change too many answers. On Qwen,
it harms 140 CS examples in MC and 202 in QA while
repairing only five and eleven CS examples. On LLaVA, it harms 91 and 192 CS
examples while repairing three and nine. Fitting on both CF and CS but
selecting settings from CF alone
causes less CS harm. It still moves LLaVA from +6.96/-0.43 to +12.17/-1.30 on
MC and from +23.91/-4.78 to +30.00/-9.57 on QA. Paired data therefore serve
two roles. Both sides shape the model for correction strength. CS outcomes limit the
cost of CF repair.

\subsection{Candidate Scoring Does Not Explain the Gain}

The preceding tests change the source of the prior scores and the data used
for fitting. We next check whether the way candidate scores are computed can
explain the gain.

The teacher-forced score used throughout the paper is
\begin{equation}
s(y)=\sum_{t=1}^{|y|}\log p(y_t\mid I,q,y_{<t}),
\end{equation}
with $I$ removed in the text-only scoring condition. The scored text is the
original answer token (yes/no or A/B/C/D). It is not the answer description. Table
\ref{tab:score-normalization} compares every answer in all 1,200 examples per
fully evaluated model.

\begin{table}[t]
\centering
\scriptsize
\setlength{\tabcolsep}{1.0pt}
\begin{tabular*}{\columnwidth}{@{\extracolsep{\fill}}lrrrrr@{}}
\toprule
Task & \shortstack{Summed-score\\$\Delta_{\mathrm{CF}}/\Delta_{\mathrm{CS}}$} &
\shortstack{Mean-score\\$\Delta_{\mathrm{CF}}/\Delta_{\mathrm{CS}}$} &
\shortstack{Agreement at\\fixed strength} &
\shortstack{Selected CF\\agreement} &
\shortstack{Selected CS\\agreement} \\
\midrule
\multicolumn{6}{l}{\itshape Qwen3-VL-32B} \\
MC & +7.83/+0.43 & +7.83/+0.43 & 100.0 & 100.0 & 100.0 \\
QA & +22.17/-1.74 & +22.17/-1.74 & 100.0 & 100.0 & 100.0 \\
\midrule
\multicolumn{6}{l}{\itshape LLaVA-1.6-34B} \\
MC & +6.96/-0.43 & +4.35/+0.87 & 100.0 & 96.1 & 97.4 \\
QA & +23.91/-4.78 & +23.04/-4.78 & 100.0 & 99.1 & 100.0 \\
\bottomrule
\end{tabular*}
\caption{\textbf{Comparison of summed and mean token log-probabilities.}
The three agreement columns compare the predictions from summed and mean
scores. They cover a shared fixed correction strength and the CF and CS
predictions after separate setting selection. All agreement values are
percentages. The two $\Delta$ columns report percentage points.}
\label{tab:score-normalization}
\end{table}

No example contains answers of unequal token length. Image-conditioned and
text-only token counts always match. Stored sums equal token count times stored
means. Answer length therefore cannot favor one candidate within an example.
Using mean scores changes the scale and spread of scores across examples.
It can therefore change which LLaVA setting is selected. Both selected
settings still give positive MC and QA CF gains. The correction direction is
unchanged because sum and mean scores agree at every fixed correction strength.

Answer length is not the only possible scoring concern. Scoring the supplied
answers could also replace the original decoder with a better classifier.
Table~\ref{tab:native-score-agreement} shows that this does not explain the
result. It compares the original answer with the top image-conditioned
candidate before prior subtraction. Their agreement is high in every model
and task combination except Qwen QA on CF images.
The top image-conditioned answer is already correct in only five of the 148 CF
repairs made by SPC across the two models and formats. In 143 repairs, the
original prediction, image-conditioned scores, and text-only scores initially
favor the same wrong answer. SPC repairs these errors by changing the candidate margin.
It does not simply select a correct answer already exposed by
image-conditioned scoring.

\begin{table}[t]
\centering
\small
\setlength{\tabcolsep}{2.0pt}
\resizebox{\columnwidth}{!}{%
\begin{tabular}{lrrrr}
\toprule
Model and task & \shortstack{Original=image-scored answer\\CF/CS (\%)} &
\shortstack{CF\\repairs} & \shortstack{Image already\\correct (\%)} &
\shortstack{All three chose\\the same answer (\%)} \\
\midrule
Qwen3-VL-32B MC & 92.2/97.8 & 21 & 19.0 & 81.0 \\
Qwen3-VL-32B QA & 75.7/99.6 & 53 & 0.0 & 100.0 \\
LLaVA-1.6-34B MC & 97.0/97.8 & 17 & 5.9 & 94.1 \\
LLaVA-1.6-34B QA & 96.5/98.3 & 57 & 0.0 & 100.0 \\
\bottomrule
\end{tabular}
}
\caption{\textbf{Agreement between the original prediction and
teacher-forced candidate scoring.} Results use
$\epsilon_{\mathrm{CS}}=0.04$. ``Image already correct'' and ``all three chose
the same answer'' are fractions of SPC's CF repairs. The latter compares the
original, image-conditioned, and text-only top answers before correction.}
\label{tab:native-score-agreement}
\end{table}

\subsection{Correction Strength and Selective Answer Revision}

Prior subtraction determines the corrected proposal. Selective answer
revision determines whether that proposal replaces the original answer.
We first vary the allowed CS decrease to trace this repair--retention balance.

\begin{figure}[t]
\centering
\includegraphics[width=\columnwidth]{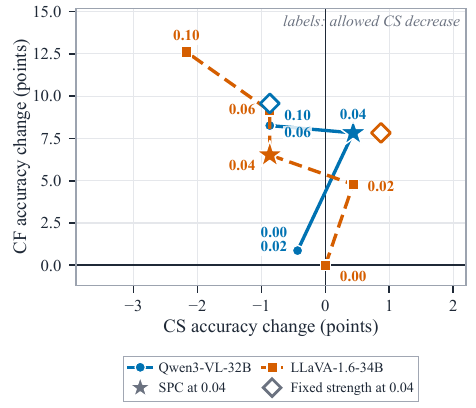}
\caption{\textbf{MC repair and retention as $\epsilon_{\mathrm{CS}}$ changes.}
The curves use the SPC version that accounts for uncertainty in the learned
correction strength. Point labels give $\epsilon_{\mathrm{CS}}$. Stars mark
$\epsilon_{\mathrm{CS}}=0.04$. Open diamonds use one fixed correction strength
selected under the same limit.}
\label{fig:supp-mc-frontier}
\end{figure}

Figure~\ref{fig:supp-mc-frontier} shows the MC repair--retention curve. On
Qwen, $\epsilon_{\mathrm{CS}}=0.04$ gives
$+7.8/+0.4$ CF/CS points. Increasing it to 0.06 gives $+8.3/-0.9$, and 0.10
selects the same result. Most of the Qwen MC gain therefore appears by 0.04.
On LLaVA, repair rises more steadily. It changes from $0.0/0.0$ at 0.00 to
$+12.6/-2.2$ at 0.10.

The open diamonds compare one fixed correction strength at
$\epsilon_{\mathrm{CS}}=0.04$. The fixed result is competitive on this full
development split. It gives more Qwen CF repair but less CS retention than the
learned correction strength. It improves both test changes for LLaVA at this
value. An instance-dependent strength does not always perform best on the full
development split. Table~\ref{tab:matched-shift} shows its clearer benefit for LLaVA
when development data are smaller or exclude the tested category.

\begin{figure*}[t]
\centering
\includegraphics[width=\textwidth]{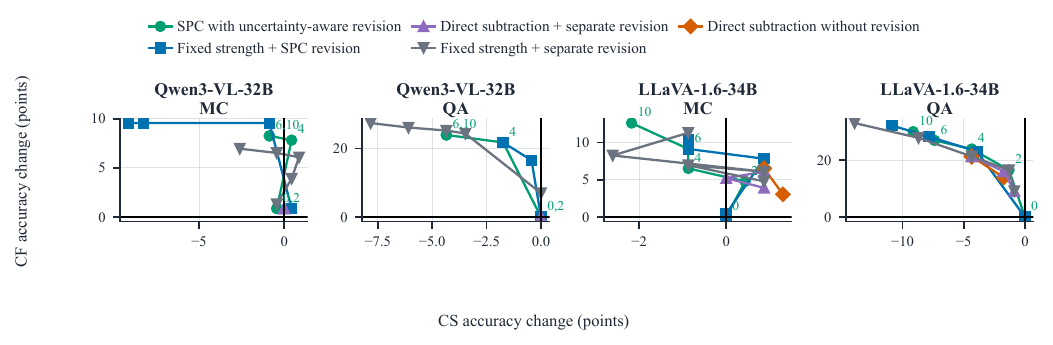}
\caption{\textbf{CF repair and CS retention across allowed CS decreases.}
Results use the uncertainty-aware extension of SPC and
$\epsilon_{\mathrm{CS}}\in\{0,0.02,0.04,0.06,0.10\}$. SPC labels show the
same limits in percentage points: 0, 2, 4, 6, and 10. Labels are combined when
settings produce the same point. Fixed-strength controls use one correction
strength. Controls based on direct subtraction remove text-only scores directly. The
two separate answer-revision controls use the same 13 features and 70
development pairs.}
\label{fig:supp-gate-pareto}
\end{figure*}

Figure~\ref{fig:supp-gate-pareto} extends the focused MC curve to QA and adds
three proposal and answer-revision controls. It shows the full balance between
CF repair and CS retention as $\epsilon_{\mathrm{CS}}$ changes. We then hold
$\epsilon_{\mathrm{CS}}=0.04$ fixed. Table~\ref{tab:matched-gate-four}
compares different ways to form and accept a corrected proposal.

The controls separate the correction strength from the final revision rule.
The fixed-strength control selects one strength on development data and keeps
SPC's selective answer revision. The learned-revision control instead fits a
logistic revision rule from the same 13 score features. We apply this rule to
proposals formed by either a fixed correction strength or direct subtraction
of the text-only scores. All controls use the same development examples,
selection objective, and value of $\epsilon_{\mathrm{CS}}$ as SPC.
We search fixed strengths in
$\{0.1,0.2,0.3,0.5,0.8,1.0,1.2,1.5\}$ and logistic-rule $\ell_2$
penalties in $\{0.01,0.1,1,10\}$. The development set also selects the
probability threshold. The selected settings (proposal scale, $\ell_2$,
threshold) are $(1.2,1,0.469)$ and $(1.1,0.01,0.532)$ for Qwen's fixed and
direct proposals. They are $(0.5,1,0.162)$ and $(2.0,1,0.162)$ for LLaVA.

\begin{table}[t]
\centering
\small
\setlength{\tabcolsep}{3.0pt}
\begin{tabular}{@{}L{0.54\columnwidth}C{0.19\columnwidth}C{0.19\columnwidth}@{}}
\toprule
Method & MC $\Delta_{\mathrm{CF}}/\Delta_{\mathrm{CS}}$ &
QA $\Delta_{\mathrm{CF}}/\Delta_{\mathrm{CS}}$ \\
\midrule
\multicolumn{3}{l}{\itshape Qwen3-VL-32B} \\
NoLan & +0.0/-4.8 & +13.9/-0.9 \\
\shortstack[l]{Adaptive strength; always\\adopt proposal} & +9.6/-2.6 & +22.6/-4.8 \\
SPC & +7.8/+0.4 & +22.2/-1.7 \\
\shortstack[l]{Fixed $\lambda=0.8$; always\\adopt proposal} & +10.4/-2.2 & +17.4/-0.9 \\
\shortstack[l]{Selected fixed strength;\\SPC revision rule} & +9.6/-0.9 & +16.5/-0.4 \\
\shortstack[l]{Fixed strength; learned\\revision rule} & +6.1/+0.9 & +25.2/-4.3 \\
\shortstack[l]{Direct subtraction; learned\\revision rule} & +0.9/+0.0 & +0.0/+0.0 \\
\midrule
\multicolumn{3}{l}{\itshape LLaVA-1.6-34B} \\
NoLan & +9.6/-2.6 & +24.8/-8.7 \\
\shortstack[l]{Adaptive strength; always\\adopt proposal} & +10.9/-0.9 & +30.9/-10.4 \\
SPC & +7.0/-0.4 & +23.9/-4.8 \\
\shortstack[l]{Fixed $\lambda=0.8$; always\\adopt proposal} & +11.3/-0.9 & +33.0/-13.9 \\
\shortstack[l]{Selected fixed strength;\\SPC revision rule} & +7.8/+0.9 & +23.0/-3.9 \\
\shortstack[l]{Fixed strength; learned\\revision rule} & +6.1/+0.9 & +21.3/-4.3 \\
\shortstack[l]{Direct subtraction; learned\\revision rule} & +6.1/+0.9 & +21.3/-4.3 \\
\bottomrule
\end{tabular}
\caption{\textbf{Effect of correction strength and selective answer revision.}
Results use $\epsilon_{\mathrm{CS}}=0.04$. Each result is
$\Delta_{\mathrm{CF}}/\Delta_{\mathrm{CS}}$ in percentage points relative to
the common original predictions. NoLan is the released dynamic
token-level rule with $\beta=0.8$. The rows differ in how they form a proposal and whether they
always adopt it, use SPC's revision rule, or use a separately learned revision
rule.}
\label{tab:matched-gate-four}
\end{table}

Table~\ref{tab:matched-gate-four} separates the corrected proposal from
selective answer revision. Always adopting the proposal increases repair and
also increases CS loss, especially on LLaVA QA. A fixed correction strength
shifts the balance between MC and QA. A separately trained rule for answer
revision allows more Qwen QA repair at a larger CS cost. NoLan improves CF in
three settings and leaves Qwen MC unchanged, but its CS cost varies across
models and answer formats.

The fixed $\lambda=0.8$ row isolates the effect of using one correction
strength without selective answer revision. It gives substantial CF repair,
but always adopting its proposal causes a 13.9-point CS loss on LLaVA QA. By
contrast, the selected fixed strength retains SPC's revision rule. The first
component row, ``Adaptive strength; always adopt proposal,'' tests correction
without selective revision. The row ``Selected fixed strength; SPC revision
rule'' tests whether one shared strength can replace SPC's
instance-dependent strength. For LLaVA, the fixed-strength and
direct-subtraction proposals agree on every development example and all but
one test example. The learned revision rule rejects that one difference, so
the two LLaVA rows have the same final results.

\subsection{Performance with Less or Shifted Development Data}

We next repeat this comparison with less development data or with the tested
CDH category excluded from development.

\begin{table}[t]
\centering
\small
\setlength{\tabcolsep}{3.4pt}
\begin{tabular*}{\columnwidth}{@{\extracolsep{\fill}}lrrr@{}}
\toprule
Development data & SPC & \shortstack{Fixed\\strength} &
\shortstack{Learned\\revision rule} \\
\midrule
\multicolumn{4}{l}{\itshape Qwen3-VL-32B} \\
Target category excluded & 11.52 & 9.89 & \textbf{14.89} \\
14 pairs & 10.72 & 9.41 & \textbf{11.83} \\
\midrule
\multicolumn{4}{l}{\itshape LLaVA-1.6-34B} \\
Target category excluded & \textbf{15.43} & 14.35 & 12.72 \\
14 pairs & \textbf{13.91} & 10.70 & 11.54 \\
\bottomrule
\end{tabular*}
\caption{\textbf{Average MC/QA utility $U_{0.5}$ in percentage points with
less or shifted development data.} Development data either exclude the tested CDH category or
contain only 14 pairs. The 14-pair result averages five samples that preserve
the category proportions. The SPC column uses the uncertainty-aware
extension.}
\label{tab:matched-shift}
\end{table}

Table~\ref{tab:matched-shift} asks whether these alternatives retain their
ordering when fewer development examples are available or one CDH category
is excluded. The separately learned revision rule has the highest average $U_{0.5}$
in the two Qwen settings. SPC is best in both LLaVA settings. Candidate-specific
prior subtraction transfers in all four conditions. The best form of answer
revision differs by model.

The selected correction strength varies across examples. For Qwen, the middle
50\% ranges from 0.236 to 0.827 on MC and from 0.395 to 1.000 on QA. For
LLaVA, it ranges from 0.713 to 0.800 and from 0.550 to 0.735. The LLaVA MC
strength reaches its maximum of 0.8 on 62.6\% of examples. This explains why a
fixed correction strength is competitive for LLaVA MC. Only 8.9\% of Qwen
MC examples reach their maximum. Table~\ref{tab:matched-shift} contains four
results that either exclude a CDH category or use 14 pairs. SPC exceeds the
fixed control by 1.08--3.21 utility points in all four tests. The learned
revision rule is stronger on Qwen, while SPC is stronger on LLaVA. These results
support adapting the correction strength, but the best final answer rule still
depends on the model.

\subsection{Contribution of the Two Support Conditions}

Removing individual conditions from selective answer revision tests the two
cases defined in the main paper. In the Qwen analysis with
$\epsilon_{\mathrm{CS}}=0.10$, removing the conflict condition costs 1.7
additional MC CS points without improving QA. Allowing changes only when the
image-conditioned and text-only scores prefer different answers produces at
most +0.9 CF points. The case in which all scores initially prefer the same
answer retains most of the gain. Many repairs therefore begin with agreement
on the original wrong answer. Prior subtraction then makes a supported
alternative the top answer. Direct disagreement between the two scoring conditions adds a
smaller set of repairs.

\subsection{Effect of Development-Set Size}

The component tests above use all 70 development pairs. We next reduce this
set while keeping the same 230 test pairs.

\begin{table}[t]
\centering
\scriptsize
\setlength{\tabcolsep}{1.2pt}
\begin{tabular*}{\columnwidth}{@{\extracolsep{\fill}}lrrrr@{}}
\toprule
\shortstack{Development\\pairs} & MC $\Delta_{\mathrm{CF}}$ & MC $\Delta_{\mathrm{CS}}$ &
QA $\Delta_{\mathrm{CF}}$ & QA $\Delta_{\mathrm{CS}}$ \\
\midrule
\multicolumn{5}{l}{\itshape Qwen3-VL-32B} \\
14 & $+5.7\pm2.5$ & $-1.1\pm2.2$ & $+22.0\pm5.1$ & $-3.5\pm2.9$ \\
28 & $+5.7\pm2.2$ & $-2.7\pm2.0$ & $+24.1\pm3.2$ & $-4.8\pm2.7$ \\
42 & $+7.8\pm1.2$ & $-1.1\pm1.6$ & $+24.4\pm1.6$ & $-4.2\pm1.6$ \\
56 & $+8.0\pm0.5$ & $-0.3\pm0.9$ & $+23.7\pm2.2$ & $-3.9\pm1.2$ \\
70 & $+8.3\pm0.0$ & $-0.9\pm0.0$ & $+23.9\pm0.0$ & $-4.3\pm0.0$ \\
\midrule
\multicolumn{5}{l}{\itshape LLaVA-1.6-34B} \\
14 & $+4.6\pm1.0$ & $-1.0\pm0.7$ & $+26.7\pm2.6$ & $-7.0\pm1.9$ \\
28 & $+10.1\pm1.8$ & $-2.4\pm1.3$ & $+29.7\pm2.1$ & $-10.5\pm2.8$ \\
42 & $+10.0\pm1.3$ & $-1.9\pm0.8$ & $+29.5\pm2.3$ & $-10.1\pm2.9$ \\
56 & $+10.4\pm1.6$ & $-1.9\pm0.7$ & $+28.3\pm2.7$ & $-8.6\pm2.7$ \\
70 & $+12.6\pm0.0$ & $-2.2\pm0.0$ & $+30.0\pm0.0$ & $-9.1\pm0.0$ \\
\bottomrule
\end{tabular*}
\caption{\textbf{Effect of development-set size with
$\epsilon_{\mathrm{CS}}=0.10$.} Every row uses the same 230-pair test set. For
each of five seeds, smaller sets are contained in larger sets. They contain one
to five pairs from every subcategory. Cells report mean $\pm$ standard
deviation in points. The 70-pair set is identical across seeds. These results
use the uncertainty-aware extension of SPC.}
\label{tab:calibration-size}
\end{table}

Table~\ref{tab:calibration-size} shows that even one pair per subcategory
produces positive mean CF change in both formats
and models. More development data mainly strengthen MC and reduce Qwen MC
variation. QA produces substantial CF gains with fewer examples, but its CS
cost differs by model. The table shows that candidate-specific prior
scores are useful early. More data are needed to stabilize answer revision.

\subsection{Fitting One SPC Model for MC and QA}

The primary SPC model is fitted jointly on MC and QA. To test the effect of
this choice, we fit and select a second model using only development MC.
Table~\ref{tab:joint-task-fitting} compares the two fits.

\begin{table}[t]
\centering
\scriptsize
\setlength{\tabcolsep}{1.5pt}
\begin{tabular*}{\columnwidth}{@{\extracolsep{\fill}}lrrrr@{}}
\toprule
& \multicolumn{2}{c}{Qwen3-VL-32B}
& \multicolumn{2}{c}{LLaVA-1.6-34B} \\
\cmidrule(lr){2-3}\cmidrule(lr){4-5}
Fitting data & MC & QA & MC & QA \\
\midrule
MC and QA & +7.8/+0.4 & +22.2/-1.7 & +7.0/-0.4 & +23.9/-4.8 \\
MC only & +9.1/-0.9 & +0.0/+0.0 & +11.3/-2.6 & +0.0/+0.0 \\
\bottomrule
\end{tabular*}
\caption{\textbf{Effect of fitting one SPC model for both answer formats.}
Each entry is $\Delta_{\mathrm{CF}}/\Delta_{\mathrm{CS}}$ in percentage
points. Both rows use $\epsilon_{\mathrm{CS}}=0.04$ on the development tasks
available during fitting.}
\label{tab:joint-task-fitting}
\end{table}

The MC-only model gives more MC repair but changes no QA answer. Joint fitting
therefore trades some MC repair for a correction that applies to both answer
formats.

\section{RQ4: Generalization of SPC}

The remaining tests change CDH categories, candidate-answer orders, data
splits, and evaluation benchmarks. All settings remain fixed after selection
on development data.

\subsection{Generalization Across CDH Categories}

For each test, we remove one CDH category from the development set.
We then evaluate only the test pairs from that category. Combining the
predictions from the three held-out-category evaluations covers all 230 test
pairs. Qwen obtains +8.7/+1.3 points on MC and +24.8/-4.3 on QA. LLaVA
obtains +8.7/-0.9 and +27.8/-6.1. Each held-out category improves in both
formats.

We also repeat this procedure for each of the 14 subcategories. Aggregating the
predictions from the 14 held-out-subcategory evaluations gives +7.0/-0.4 on
Qwen MC and +24.3/-3.9 on Qwen QA. MC improves in nine subcategories, stays
unchanged in three, and declines in two. QA improves in 13 and stays unchanged
in one. No category or subcategory label enters the feature vector or setting
selection.

\begin{table}[t]
\centering
\scriptsize
\setlength{\tabcolsep}{1.0pt}
\begin{tabular}{@{}L{0.20\columnwidth}C{0.165\columnwidth}C{0.18\columnwidth}
                C{0.165\columnwidth}C{0.18\columnwidth}@{}}
\toprule
\shortstack[l]{Held-out\\category} &
\shortstack{MC\\$\Delta_{\mathrm{CF}}/\Delta_{\mathrm{CS}}$} &
\shortstack{MC\\repairs/harms\\(CF; CS)} &
\shortstack{QA\\$\Delta_{\mathrm{CF}}/\Delta_{\mathrm{CS}}$} &
\shortstack{QA\\repairs/harms\\(CF; CS)} \\
\midrule
\multicolumn{5}{l}{\itshape Qwen3-VL-32B} \\
Attribute & +9.3/+0.0 & 7/0;1/1 & +10.7/-9.3 & 10/2;0/7 \\
Counting & +10.0/+0.0 & 8/0;1/1 & +33.8/-3.8 & 27/0;0/3 \\
Relation & +6.7/+4.0 & 7/2;3/0 & +29.3/+0.0 & 22/0;0/0 \\
\midrule
\multicolumn{5}{l}{\itshape LLaVA-1.6-34B} \\
Attribute & +5.3/+0.0 & 4/0;0/0 & +5.3/-1.3 & 6/2;0/1 \\
Counting & +10.0/-1.2 & 8/0;0/1 & +41.2/-5.0 & 33/0;0/4 \\
Relation & +10.7/-1.3 & 10/2;3/4 & +36.0/-12.0 & 27/0;0/9 \\
\midrule
\multicolumn{5}{l}{\itshape Qwen3-VL-30B-A3B} \\
Attribute & +9.3/-5.3 & 7/0;1/5 & -- & -- \\
Counting & +8.8/-2.5 & 7/0;0/2 & -- & -- \\
Relation & +6.7/+1.3 & 7/2;3/2 & -- & -- \\
\bottomrule
\end{tabular}
\caption{\textbf{Generalization to CDH categories excluded from fitting.}
Results are in percentage points. Within a
repairs/harms cell, the values before the semicolon describe CF and those
after it describe CS. These results use the uncertainty-aware extension of
SPC.}
\label{tab:family-folds}
\end{table}

The additional model also shows directed prior attraction. It makes 84
original MC errors on CF images. Of these, 68 (81.0\%) select the
prior-favored candidate. That candidate is the paired CS answer in 76 cases
(90.5\%). On all 230 test pairs, SPC changes CF/CS accuracy by +9.6/-4.3
points with 24/2 CF repairs/harms. Every excluded CDH category retains a
positive CF change in Table~\ref{tab:family-folds}. This gives the same
qualitative conclusion as the two main models.

\subsection{Stability Across Data Splits}

\begin{figure}[t]
\centering
\includegraphics[width=\columnwidth]{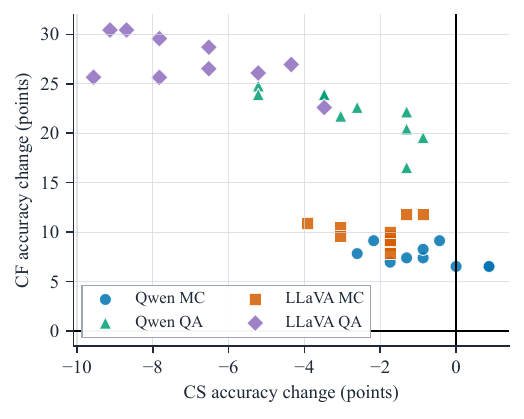}
\caption{\textbf{Results over ten independent 70/230 data splits.} Each split
keeps five development pairs from every subcategory, and SPC is fitted again.
Every point has positive CF change. CS change varies across splits. These
results use the uncertainty-aware extension of SPC.}
\label{fig:repeated-splits}
\end{figure}

Figure~\ref{fig:repeated-splits} shows the ten independent fits.
For each of ten predefined seeds, we sample five pairs from every
subcategory. Both image sides and both tasks stay in the same split. SPC is
then fitted on 70 pairs and tested on the other 230.
Table~\ref{tab:repeated-split-summary} reports the corresponding means and
ranges.

\begin{table}[t]
\centering
\scriptsize
\setlength{\tabcolsep}{2.0pt}
\begin{tabular*}{\columnwidth}{@{\extracolsep{\fill}}lrr@{}}
\toprule
Model and task & \shortstack{$\Delta_{\mathrm{CF}}$\\mean $\pm$ std. [range]} &
\shortstack{$\Delta_{\mathrm{CS}}$\\mean $\pm$ std. [range]} \\
\midrule
Qwen MC & $+7.57\pm1.01$ [6.52, 9.13] & $-0.83\pm1.19$ [-2.61, 0.87] \\
Qwen QA & $+21.96\pm2.53$ [16.52, 24.78] & $-2.78\pm1.60$ [-5.22, -0.87] \\
LLaVA MC & $+9.96\pm1.27$ [7.83, 11.74] & $-2.09\pm0.93$ [-3.91, -0.87] \\
LLaVA QA & $+27.26\pm2.50$ [22.61, 30.43] & $-6.91\pm2.06$ [-9.57, -3.48] \\
\bottomrule
\end{tabular*}
\caption{\textbf{Variation across ten independent data splits.} All values are
percentage points. All 40 CF changes are positive.}
\label{tab:repeated-split-summary}
\end{table}

\subsection{Generalization Across MC Candidate-Answer Orders}

Category transfer keeps the original answer order. The next test keeps the
answer meanings fixed but rotates or reverses their positions. Rotation 1
changes the old order A/B/C/D to B/C/D/A. Rotation 2 changes it to C/D/A/B.
Reversal changes it to D/C/B/A. The answer letters are updated after each
change. Table~\ref{tab:mc-option-order} reports the resulting accuracy.

\begin{table}[t]
\centering
\scriptsize
\setlength{\tabcolsep}{1.8pt}
\begin{tabular*}{\columnwidth}{@{\extracolsep{\fill}}lrrr@{}}
\toprule
\shortstack{Candidate-answer\\order} & \shortstack{Original\\CF/CS Acc.} &
\shortstack{SPC\\CF/CS Acc.} &
$\Delta_{\mathrm{CF}}/\Delta_{\mathrm{CS}}$ \\
\midrule
\multicolumn{4}{l}{\itshape Qwen3-VL-32B} \\
Original & 0.665/0.965 & 0.748/0.957 & +8.3/-0.9 \\
Rotation 1 & 0.665/0.961 & 0.739/0.939 & +7.4/-2.2 \\
Rotation 2 & 0.678/0.961 & 0.757/0.922 & +7.8/-3.9 \\
Reversal & 0.687/0.970 & 0.739/0.939 & +5.2/-3.0 \\
\midrule
\multicolumn{4}{l}{\itshape LLaVA-1.6-34B} \\
Original & 0.565/0.917 & 0.691/0.896 & +12.6/-2.2 \\
Rotation 1 & 0.561/0.939 & 0.687/0.917 & +12.6/-2.2 \\
Rotation 2 & 0.530/0.943 & 0.670/0.891 & +13.9/-5.2 \\
Reversal & 0.539/0.922 & 0.674/0.878 & +13.5/-4.3 \\
\bottomrule
\end{tabular*}
\caption{\textbf{MC results after changing candidate-answer positions.} SPC uses
$\epsilon_{\mathrm{CS}}=0.10$ on the original development data. Answer texts
and meanings are unchanged. The final column reports percentage points. These
results use the uncertainty-aware extension of SPC.}
\label{tab:mc-option-order}
\end{table}

With $\epsilon_{\mathrm{CS}}=0.04$, Qwen's CF changes range from +5.7 to
+7.8 points. Its CS changes range from +0.4 to -3.0. LLaVA's CF changes range
from +6.5 to +10.0, and its CS changes range from -0.9 to -2.6. With
$\epsilon_{\mathrm{CS}}=0.10$, the largest CF accuracy difference across
orders is 1.7 points for Qwen and 2.2 points for LLaVA.

Exact paired tests cover 16 comparisons. These include two models, four
candidate-answer orders, and two image sides. Seven of eight CF gains remain
significant at 0.05. Only Qwen reversal does not (Holm $p=0.136$). For CS,
only LLaVA Rotation 2 remains significant (Holm $p=0.0165$).

The four orders also reveal a limitation. We first map answer letters back to
their meanings. The original predictions then agree across all four orders on
86.5/95.2\% of Qwen CF/CS pairs and 85.2/91.3\% of LLaVA pairs. SPC with
$\epsilon_{\mathrm{CS}}=0.10$ obtains 78.7/90.9\% and 78.3/84.8\%. SPC
therefore does not always produce the same answer after candidate positions
change. Still, CF accuracy varies little and every paired CF change is
positive. The gain therefore survives changes in candidate-answer order.

\subsection{Transfer to Other Hallucination Benchmarks}

The external benchmarks use different structures. Visual CounterFact provides
matched ordinary and counterfactual images
\cite{golovanevsky2025pixels}. HallusionBench pairs easy and hard questions
\cite{guan2024hallusionbench}. We treat the hard question as the conflict side
and its matched easy question as the control side. ConflictVIS contains MC and binary QA
\cite{liu2025insight}. POPE and POPEv2 test object presence rather than a
matched ordinary-state answer \cite{li2023pope,li2026popev2}. We report each
benchmark in its native subsets and metrics. Open-ended benchmarks such as
CHAIR and MMHal-Bench are outside SPC's current fixed-candidate setting
\cite{rohrbach2018chair,sun2024mmhal}.

Table~\ref{tab:supp-transfer} shows transfer with settings selected only
on CDH development data. The paragraphs that follow give the protocol and
analysis for each benchmark.

\begin{table}[t]
\centering
\scriptsize
\setlength{\tabcolsep}{1.4pt}
\resizebox{\columnwidth}{!}{%
\begin{tabular}{llrrrrr}
\toprule
Benchmark & Subset & Original metric &
\shortstack{SPC metric\\$\epsilon_{\mathrm{CS}}=0.04$} &
\shortstack{SPC metric\\$\epsilon_{\mathrm{CS}}=0.10$}
& \shortstack{Repairs/harms\\$\epsilon_{\mathrm{CS}}=0.04$} &
\shortstack{Repairs/harms\\$\epsilon_{\mathrm{CS}}=0.10$} \\
\midrule
Visual CounterFact (Qwen) & CF & 0.962 & 0.993 & 0.997 & 38/0 & 42/0 \\
 & CS & 0.998 & 0.995 & 0.989 & 2/5 & 2/12 \\
Visual CounterFact (LLaVA) & CF & 0.939 & 0.984 & 0.989 & 55/0 & 62/1 \\
 & CS & 0.994 & 0.993 & 0.995 & 1/2 & 3/2 \\
HallusionBench & Hard & 0.607 & 0.659 & 0.668 & 20/3 & 23/3 \\
 & Easy & 0.848 & 0.838 & 0.820 & 4/7 & 5/14 \\
ConflictVIS & MC & 0.971 & 0.992 & 0.992 & 8/0 & 8/0 \\
 & QA & 0.738 & 0.995 & 0.995 & 96/0 & 96/0 \\
POPE & All & 0.899/0.893 & 0.898/0.893 & 0.898/0.893 & 6/11 & 6/12 \\
POPEv2 & Absent & 0.686 & 0.671 & 0.671 & 0/1 & 0/1 \\
 & Present & 0.986 & 0.986 & 0.986 & 0/0 & 0/0 \\
\bottomrule
\end{tabular}
}
\caption{\textbf{Performance on external benchmarks.} SPC settings are selected
on CDH development data with $\epsilon_{\mathrm{CS}}=0.04$ or $0.10$.
POPE reports accuracy/F1. All other entries report accuracy.
Repairs/harms are measured against the same original predictions. Visual
CounterFact is evaluated with both listed models; the remaining benchmarks use
Qwen3-VL-32B.}
\label{tab:supp-transfer}
\end{table}

With $\epsilon_{\mathrm{CS}}=0.10$, the CF gains remain significant on Visual
CounterFact ($p=4.5\times10^{-13}$) and HallusionBench Hard
($p=8.8\times10^{-5}$). Both ConflictVIS gains also remain significant
($p=0.0078$ and $p=2.5\times10^{-29}$). The larger Qwen Visual CounterFact CS
change is significant ($p=0.0129$). HallusionBench Easy is not statistically
significant ($p=0.0636$). POPE is also not statistically significant
($p=0.238$). Neither POPEv2 side changes significantly ($p=1.0$ for both).
The 0.10 setting repairs more conflict cases. The
0.04 setting better preserves the matched CS side.

We also evaluate NoLan with its published $\beta=0.8$ rule. Relative to the
same original predictions used for SPC, its Qwen changes are +3.1/-0.7 on
Visual CounterFact and -2.1/-7.6 on HallusionBench. They are +1.3 and +24.1
on ConflictVIS MC and QA. On POPE, accuracy/F1 changes by +0.0/+0.6. On
POPEv2, absent/present accuracy changes by -21.4/+0.0. The answer-token
implementation uses full-vocabulary logits. Every value here uses the common
original predictions as its reference.

Visual CounterFact \cite{golovanevsky2025pixels} provides a large paired
transfer test. We use all 1,220 released color and size pairs. We retain the
published questions and A/B balance. Original images form CS, and the
published counterfactual images form CF. SPC settings come only from the 70 CDH
development pairs.

The text-only scores rank the ordinary attribute first on 92.8\% of Qwen
pairs and 82.1\% of LLaVA pairs. The prior-favored candidate matches 45 of
Qwen's 46 original CF errors and all 75 LLaVA errors. All 121 errors select
the ordinary attribute.

With $\epsilon_{\mathrm{CS}}=0.04$, Qwen changes CF/CS by +3.11/-0.25
points. LLaVA changes them by +4.51/-0.08. The Qwen CF interval is
[2.2, 4.1] points with 38/0 repairs/harms. The LLaVA interval is
[3.4, 5.7] points with 55/0. Exact CF tests give
$p=7.28\!\times\!10^{-12}$ and $5.55\!\times\!10^{-17}$. The CS changes are
not significant.

For Qwen, the color and size changes are +1.01/-0.81 and +4.54/+0.14. For
LLaVA, they are +4.87/-0.41 and +4.26/+0.14. SPC therefore transfers beyond
the CDH categories and benchmark used for fitting. Results with
$\epsilon_{\mathrm{CS}}=0.10$ appear in Table~\ref{tab:vcf-controls}.

\begin{table}[t]
\centering
\small
\setlength{\tabcolsep}{2.0pt}
\begin{tabular}{@{}L{0.25\columnwidth}L{0.39\columnwidth}
                C{0.14\columnwidth}C{0.14\columnwidth}@{}}
\toprule
Model & Method selected on CDH & $\Delta_{\mathrm{CF}}$ &
$\Delta_{\mathrm{CS}}$ \\
\midrule
Qwen3-VL-32B & SPC ($\epsilon_{\mathrm{CS}}=0.04$) & +3.11 & -0.25 \\
 & SPC ($\epsilon_{\mathrm{CS}}=0.10$) & +3.44 & -0.82 \\
 & \shortstack[l]{Fixed correction strength\\($\epsilon_{\mathrm{CS}}=0.10$)} & +2.38 & -0.25 \\
 & \shortstack[l]{Separate answer revision\\($\epsilon_{\mathrm{CS}}=0.10$)} & +3.44 & -4.02 \\
LLaVA-1.6-34B & SPC ($\epsilon_{\mathrm{CS}}=0.04$) & +4.51 & -0.08 \\
 & SPC ($\epsilon_{\mathrm{CS}}=0.10$) & +5.00 & +0.08 \\
 & \shortstack[l]{Fixed correction strength\\($\epsilon_{\mathrm{CS}}=0.10$)} & +3.93 & -0.16 \\
 & \shortstack[l]{Separate answer revision\\($\epsilon_{\mathrm{CS}}=0.10$)} & +5.49 & -0.33 \\
\bottomrule
\end{tabular}
\caption{\textbf{Transfer to Visual CounterFact in accuracy points.} Every
variant is selected on CDH development data only.}
\label{tab:vcf-controls}
\end{table}

Table~\ref{tab:vcf-controls} separates candidate-specific prior subtraction
from correction strength and selective answer revision. Transfer remains
positive with a fixed strength and with separately learned answer revision.
SPC provides the better balance between CF repair and CS retention. Separate
answer revision matches or exceeds CF repair but incurs a much larger Qwen CS
loss.

The HallusionBench transfer \cite{guan2024hallusionbench} uses 328 questions.
Each question has an exact easy/hard pair with opposite ground truth. With
$\epsilon_{\mathrm{CS}}=0.04$, SPC changes CF/CS by +5.18/-0.91 points. CF
has 20 repairs and three harms ($p=0.00049$). CS has four repairs and seven
harms ($p=0.55$).

REVIS uses its reported Qwen correction value, $\alpha=1.6$. It changes CF/CS by
+0.61/-0.91 points. The intervals are [-1.2, 2.4] and [-2.7, 0.9] points.
Its risk gate is active on every example and at 33.7\% of hidden positions.
Both methods use the same BF16 original predictions and image processing.

POPE evaluation \cite{li2023pope} uses all 9,000 released questions.
Original prediction and candidate scoring use the same 512-by-512 image processing.
Baseline accuracy/F1 is 0.917/0.911 on random, 0.898/0.892 on popular, and
0.881/0.877 on adversarial. Across the three splits, SPC with
$\epsilon_{\mathrm{CS}}=0.04$ changes 17 answers. These changes include six
repairs and 11 harms. Overall accuracy changes by -0.06 points, and F1
changes by -0.04 points ($p=0.33$).

For POPEv2, we group the released annotations by image and retain the 70
image IDs selected after shuffling with seed 20260711. The modified image, from
which the target object was removed, forms the absent side. Its original COCO
image forms the present side. Both use the same native yes/no question. SPC
changes one absent-side question and no present-side question. Selective answer
revision therefore usually keeps the original prediction on this subset.

ConflictVIS evaluation \cite{liu2025insight} uses every released question with
complete candidate scores. It contains 374 QA and 374 MC questions over 374
images. These cover 171 action conflicts and 203 place conflicts. We exclude
the open-ended subset because SPC requires a fixed candidate set. No
ConflictVIS label is used for fitting or selection.

On MC, SPC changes accuracy from 0.971 to 0.992. This is a gain of 2.14 points
with eight repairs and no harms ($p=0.0078$). The CFG score control gains 0.53
points. On QA, SPC changes accuracy from 0.738 to 0.995, while the CFG control
reaches 0.997. These QA gains require care because all 374 labels are yes and
there is no matched CS control. The MC result is therefore the more informative
transfer test.

\section{RQ5: Scope of Prior-Based Correction}
\label{sec:supp-rq5}

Binary QA can reward a method that follows a frequent answer token instead of
the visual claim. We first measure the unequal answer directions in the
original question form. We then reverse the same claims and vary how the
question is written.

The balanced claim-order test uses the original MC answer texts for 299 pairs.
Pair 222 has identical CF and CS labels in its MC annotation, so its two
caption answers provide the two answer texts. These texts appear only in the
constructed question. They are not passed to SPC as labels or metadata. Let
$a_{\mathrm{CS}}$ and $a_{\mathrm{CF}}$ denote the mutually exclusive
ordinary and atypical answers. One question places $a_{\mathrm{CS}}$ first,
and the other places $a_{\mathrm{CF}}$ first. Across both orders, each image
has one ``yes'' label and one ``no'' label. The 70/230 pair split is unchanged.

\subsection{Unequal Yes/No Answers in the Original Questions}

In the original question form, the two CF answer groups are highly unequal.
On the 218 CF examples whose correct answer is no, SPC changes Qwen accuracy
by +26.1 points. On the 12 examples whose correct answer is yes, it changes
accuracy by -16.7 points. The corresponding LLaVA changes are +32.6 and
-16.7 points. On CS, the 218 examples
whose correct answer is yes change by -4.6 and -9.6 points for Qwen and
LLaVA. Neither model changes accuracy on the 12 examples whose correct answer
is no. Overall QA accuracy on this question form therefore cannot show whether the
method works in both yes/no answer directions.

The balanced claim-order test exchanges the two claims. It keeps the image and
visual question fixed. After fitting on the original question form, Qwen
changes ordinary-first CF by +37.8 points and atypical-first CF by +4.3. Their
95\% intervals are [31.7, 44.3] and [1.3, 7.8] points. After fitting on both
claim orders, the gains are +36.1 and +11.7. Their intervals are
[30.0, 42.6] and [7.4, 16.5] points.

For LLaVA, fitting on the original form gives +33.0 and -1.7 points. The
intervals are [27.0, 39.1] and [-6.5, 3.0] points. Fitting on both claim orders
gives +46.5 and -1.3 points. The intervals are [40.0, 53.0] and
[-6.1, 3.5] points. The smaller LLaVA direction is not statistically resolved.
Table~\ref{tab:polarity-consistency} reports the corresponding consistency and
joint accuracy across both claim orders.

\begin{table}[t]
\centering
\scriptsize
\setlength{\tabcolsep}{2.0pt}
\begin{tabular*}{\columnwidth}{@{\extracolsep{\fill}}llrr@{}}
\toprule
Method & Image type & \shortstack{Consistency after\\exchanging claims} &
\shortstack{Both claim orders\\correct} \\
\midrule
\multicolumn{4}{l}{\itshape Qwen3-VL-32B} \\
\shortstack[l]{SPC fitted on\\original form} & CF & 0.761$\rightarrow$0.817 & 0.448$\rightarrow$0.687 \\
 & CS & 0.983$\rightarrow$0.939 & 0.974$\rightarrow$0.913 \\
\shortstack[l]{SPC fitted on\\both orders} & CF & 0.761$\rightarrow$0.917 & 0.448$\rightarrow$0.765 \\
 & CS & 0.983$\rightarrow$0.943 & 0.974$\rightarrow$0.926 \\
CFG score control & CF & 0.761$\rightarrow$0.422 & 0.448$\rightarrow$0.265 \\
 & CS & 0.983$\rightarrow$0.943 & 0.974$\rightarrow$0.935 \\
\midrule
\multicolumn{4}{l}{\itshape LLaVA-1.6-34B} \\
\shortstack[l]{SPC fitted on\\original form} & CF & 0.600$\rightarrow$0.843 & 0.304$\rightarrow$0.583 \\
 & CS & 0.922$\rightarrow$0.961 & 0.917$\rightarrow$0.948 \\
\shortstack[l]{SPC fitted on\\both orders} & CF & 0.600$\rightarrow$0.800 & 0.304$\rightarrow$0.630 \\
 & CS & 0.922$\rightarrow$0.883 & 0.917$\rightarrow$0.870 \\
CFG score control & CF & 0.600$\rightarrow$0.730 & 0.304$\rightarrow$0.470 \\
 & CS & 0.922$\rightarrow$0.948 & 0.917$\rightarrow$0.939 \\
\bottomrule
\end{tabular*}
\caption{\textbf{Consistency after exchanging claims.} The second measure
requires both claim orders to be correct. Results use the 230-pair balanced
claim-order test. Every cell reports before$\rightarrow$after accuracy.}
\label{tab:polarity-consistency}
\end{table}

The original prediction and the two candidate scoring conditions show where
the difference between claim orders first appears. This analysis comes before
selective answer revision.
Table~\ref{tab:prior-direction} asks whether each condition gives
opposite predictions after the two claims are exchanged. The original
prediction changes appropriately on most CS pairs but less often on CF pairs.
The text-only prediction changes on only 58.3\% of Qwen pairs and 47.4\% of
LLaVA pairs.

\begin{table}[t]
\centering
\scriptsize
\setlength{\tabcolsep}{2.5pt}
\begin{tabular*}{\columnwidth}{@{\extracolsep{\fill}}llrr@{}}
\toprule
Model & Prediction source & CF & CS \\
\midrule
Qwen & Original answer & 0.761 & 0.983 \\
 & Image-conditioned scores & 0.378 & 0.917 \\
 & Text-only scores & 0.583 & 0.583 \\
\midrule
LLaVA & Original answer & 0.600 & 0.922 \\
 & Image-conditioned scores & 0.552 & 0.891 \\
 & Text-only scores & 0.474 & 0.474 \\
\bottomrule
\end{tabular*}
\caption{\textbf{Pre-correction consistency under claim exchange.} A
prediction is consistent when exchanging the ordinary and atypical claims also
reverses its yes/no answer for the same image.}
\label{tab:prior-direction}
\end{table}

For ordinary-first CF questions, the text-only scores favor the ordinary answer
on 99.1\% of examples for both models. When the atypical claim is placed first,
they favor the ordinary answer on only 57.4\% of Qwen examples and 46.5\% of
LLaVA examples. This explains the difference between the two claim
orders. Balanced development data improve Qwen's weaker direction. They do
not make LLaVA improve in both directions. In more than half of the LLaVA
examples, subtraction no longer moves away from the ordinary answer.

The CFG score control tests global prior suppression. On Qwen, its CF change
is -19.1 points for ordinary-first questions and +16.5 for atypical-first
questions. Accuracy requiring both claim orders to be correct falls to 0.265.
On LLaVA, the two changes are +17.4 and +2.6, and joint accuracy rises to
0.470.

Neither CFG nor SPC works reliably in both claim orders for every model. On
the Contrastive form, SPC improves both directions on Qwen. It also improves
consistency after exchanging claims for both models. The LLaVA result shows the
remaining QA failure directly.

\subsection{Transfer Across Question Forms}
\suppressfloats[t]

We next change the question form, not only the claim order. Every form presents
the same two mutually exclusive visual claims. It does not reveal the CF or CS
identity. We fit SPC on both claim orders of 70 development pairs using the
Contrastive form. We then apply the same model and thresholds to 230 test
pairs. The four forms are \emph{Contrastive}, \emph{Instead},
\emph{Choice}, and \emph{Reject second}. The last three are absent from
development. Table~\ref{tab:question-form-templates} gives the exact templates.

\begin{table}[!b]
\centering
\scriptsize
\setlength{\tabcolsep}{2.0pt}
\begin{tabular}{@{}L{0.20\columnwidth}L{0.74\columnwidth}@{}}
\toprule
Question form & Template \\
\midrule
Contrastive & For the visual question ``Q,'' is the correct answer ``FIRST''
rather than ``SECOND''? \\
Instead & Looking only at the image, should the question ``Q'' be answered
``FIRST'' instead of ``SECOND''? \\
Choice & Between ``FIRST'' and ``SECOND,'' does the image support ``FIRST'' as
the answer to ``Q''? \\
Reject second & For the image-based question ``Q,'' would choosing ``SECOND''
be incorrect because ``FIRST'' is the right answer? \\
\bottomrule
\end{tabular}
\caption{\textbf{Question forms used in the claim-order test.} FIRST and
SECOND are replaced by the two answer texts. Exchanging them creates the
ordinary-first and atypical-first versions.}
\label{tab:question-form-templates}
\end{table}

The CFG score control uses one strength selected on the same development
split. Neither method uses test labels or separate thresholds for each form.
Table~\ref{tab:template-transfer} reports transfer to all four forms.

\begin{table}[!b]
\centering
\footnotesize
\setlength{\tabcolsep}{1.5pt}
\resizebox{\columnwidth}{!}{%
\begin{tabular}{lrrrr}
\toprule
Question form & \shortstack{Text-only answer reverses\\with claim order (\%)} &
\shortstack{SPC $\Delta_{\mathrm{CF}}$\\ordinary/atypical first} &
\shortstack{CFG $\Delta_{\mathrm{CF}}$\\ordinary/atypical first} &
\shortstack{Both claim orders\\correct} \\
\midrule
\multicolumn{5}{l}{\itshape Qwen3-VL-32B} \\
Contrastive & 58.3 & +36.1/+11.7 & -19.1/+16.5 & 44.8$\rightarrow$76.5 \\
Instead & 46.5 & +17.0/+10.0 & -18.7/+19.1 & 45.7$\rightarrow$63.0 \\
Choice & 52.6 & +3.0/+10.9 & -11.3/+22.2 & 54.3$\rightarrow$63.0 \\
Reject second & 24.3 & +47.0/+1.7 & -27.8/+22.6 & 30.4$\rightarrow$65.2 \\
\midrule
\multicolumn{5}{l}{\itshape LLaVA-1.6-34B} \\
Contrastive & 47.4 & +46.5/-1.3 & +17.4/+2.6 & 30.4$\rightarrow$63.0 \\
Instead & 14.3 & +57.0/-11.3 & +18.3/-2.6 & 28.3$\rightarrow$59.1 \\
Choice & 5.7 & +21.3/-13.9 & +12.2/-10.4 & 44.3$\rightarrow$45.7 \\
Reject second & 0.0 & +82.2/-43.0 & +29.1/-23.9 & 0.9$\rightarrow$49.1 \\
\bottomrule
\end{tabular}
}
\caption{\textbf{Transfer across question forms with the same meaning.}
Settings do not change for each form. The percentage is the fraction of pairs
for which exchanging the ordinary and atypical claims also reverses the
text-only prediction. Values in the two $\Delta_{\mathrm{CF}}$ columns list
the ordinary claim first and the atypical claim first. They report percentage
points. The final column gives accuracy when both claim
orders must be correct, before and after correction.}
\label{tab:template-transfer}
\end{table}

Qwen improves CF in all eight combinations of question form and claim order. Its eight CS changes
remain between -3.9 and +3.9 points. Six CF changes remain significant after
Holm adjustment over 32 comparisons. These cover two models, four question
forms, two claim orders, and two image sides.

The unresolved ordinary-first \emph{Choice} result is +3.0 points
($p=0.092$, 95\% interval [0.0, 6.1] points). The atypical-first \emph{reject
second} result is +1.7 points ($p=0.557$, interval [-2.6, 6.1] points).
As an image-free control, we select among keeping the original answer, always
answering yes, always answering no, and reversing yes/no. Development
selection chooses the original answer. A rule based only on claim order
therefore cannot reproduce the result. The CFG control loses CF on every ordinary-first Qwen form and
gains CF on every atypical-first form.

LLaVA shows a different failure. Its text-only prediction changes with claim
order on 47.4\% of examples under the fitted form. This rate falls to 14.3\%,
5.7\%, and 0.0\% on the three unseen forms. SPC then gains on ordinary-first
questions but loses on atypical-first questions. The CFG control shows the
same split.

On \emph{Reject second}, the two SPC CF changes are +82.2 and -43.0 points.
The corresponding CS changes are -33.9 and +67.8 points. Both
candidate-specific subtraction and global guidance fail in the same direction.
The problem therefore arises in prior estimation. The text-only scores follow
syntax instead of the ordinary claim.

\subsection{When the Estimated Prior Favors the Ordinary Answer}

An overall consistency rate can hide which pairs have a useful prior estimate.
We therefore divide the test pairs before measuring accuracy. We say that the
estimated prior \emph{favors the ordinary answer in both claim orders} when two
conditions hold. Its top answer is yes for the ordinary-first question. Its top
answer is no for the atypical-first question. This definition uses only
text-only scores. It does not use the
image, CF/CS identity, or correct answer. Table
\ref{tab:semantic-conditioning-full} combines the two claim orders in each
group. The main paper reports the two directions separately.

\begin{table}[t]
\centering
\scriptsize
\setlength{\tabcolsep}{1.0pt}
\begin{tabular}{@{}L{0.22\columnwidth}C{0.08\columnwidth}C{0.12\columnwidth}
                C{0.19\columnwidth}C{0.12\columnwidth}C{0.19\columnwidth}@{}}
\toprule
Estimated prior & Pairs & $\Delta_{\mathrm{CF}}$ & 95\% CI &
$\Delta_{\mathrm{CS}}$ &
\shortstack{Repairs/harms\\(CF; CS)} \\
\midrule
\multicolumn{6}{l}{\itshape Qwen3-VL-32B: Contrastive} \\
Favors ordinary & 132 & +34.47 & [28.03,41.29] & -3.79 & 91/0;0/10 \\
\shortstack[l]{Does not favor\\ordinary} & 98 & +9.69 & [5.61,14.29] & -1.53 & 22/3;0/3 \\
\multicolumn{6}{l}{\itshape Qwen3-VL-32B: Instead} \\
Favors ordinary & 106 & +23.58 & [17.45,30.19] & -0.47 & 50/0;0/1 \\
\shortstack[l]{Does not favor\\ordinary} & 124 & +4.84 & [2.42,7.66] & +0.00 & 12/0;0/0 \\
\multicolumn{6}{l}{\itshape Qwen3-VL-32B: Choice} \\
Favors ordinary & 113 & +15.49 & [11.06,20.35] & +0.00 & 35/0;0/0 \\
\shortstack[l]{Does not favor\\ordinary} & 117 & -1.28 & [-3.42,0.43] & +0.85 & 1/4;2/0 \\
\multicolumn{6}{l}{\itshape Qwen3-VL-32B: Reject second} \\
Favors ordinary & 56 & +41.07 & [32.14,50.89] & -0.89 & 46/0;0/1 \\
\shortstack[l]{Does not favor\\ordinary} & 174 & +18.97 & [14.94,22.99] & +0.29 & 77/11;10/9 \\
\midrule
\multicolumn{6}{l}{\itshape LLaVA-1.6-34B: Contrastive} \\
Favors ordinary & 107 & +33.18 & [27.10,39.72] & -4.67 & 71/0;0/10 \\
\shortstack[l]{Does not favor\\ordinary} & 123 & +13.41 & [8.13,18.70] & -1.22 & 51/18;10/13 \\
\multicolumn{6}{l}{\itshape LLaVA-1.6-34B: Instead} \\
Favors ordinary & 33 & +34.85 & [24.24,45.45] & -1.52 & 23/0;0/1 \\
\shortstack[l]{Does not favor\\ordinary} & 197 & +20.81 & [17.01,24.62] & -4.57 & 111/29;13/31 \\
\multicolumn{6}{l}{\itshape LLaVA-1.6-34B: Choice} \\
Favors ordinary & 13 & +15.38 & [0.00,34.62] & -11.54 & 4/0;0/3 \\
\shortstack[l]{Does not favor\\ordinary} & 217 & +3.00 & [-0.23,5.99] & -2.53 & 46/33;10/21 \\
\multicolumn{6}{l}{\itshape LLaVA-1.6-34B: Reject second} \\
Favors ordinary & 0 & -- & -- & -- & 0/0;0/0 \\
\shortstack[l]{Does not favor\\ordinary} & 230 & +19.57 & [15.22,23.91] & +16.96 & 189/99;156/78 \\
\bottomrule
\end{tabular}
\caption{\textbf{Results grouped by whether the estimated prior favors the
ordinary answer in both claim orders.} The reported $\Delta$ values combine both claim orders.
They are percentage points. Confidence intervals resample image pairs. The final column gives CF
repairs/harms followed by CS repairs/harms.}
\label{tab:semantic-conditioning-full}
\end{table}

Every group with a prior estimate that favors the ordinary answer improves in
both claim orders. These groups also have zero CF harms in the combined
analysis. A group without this alignment can still have a positive combined
result when its ordinary-first gain is larger than its atypical-first loss.
For example, LLaVA \emph{Reject
second} has a combined change of +19.57 even though the two changes are +82.17
and -43.04 points. The combined result must therefore be read with both
directions. In this test, SPC is most reliable across both claim orders when
the estimated prior favors the ordinary answer. Otherwise, one direction can
dominate the average. This is a limit of the prior estimate, not a lack of
confidence in the corrected proposal. Raising a confidence threshold cannot
fix a score whose meaning changes with question form or claim order.

\section{Statistical Tests and Computational Cost}

The main results compare paired predictions before and after correction.
Table~\ref{tab:primary-holm} gives the corresponding exact tests. We then
report the resampling procedure and computational cost.

\begin{table}[t]
\centering
\scriptsize
\setlength{\tabcolsep}{3.0pt}
\begin{tabular*}{\columnwidth}{@{\extracolsep{\fill}}llrr@{}}
\toprule
Model & Comparison & Raw $p$ & Holm $p$ \\
\midrule
Qwen3-VL-32B & MC CF & 0.000277 & 0.001662 \\
 & MC CS & 1.000000 & 1.000000 \\
 & QA CF & $8.55\!\times\!10^{-14}$ & $7.70\!\times\!10^{-13}$ \\
 & QA CS & 0.125000 & 0.375000 \\
LLaVA-1.6-34B & MC CF & 0.000145 & 0.000870 \\
 & MC CS & 1.000000 & 1.000000 \\
 & QA CF & $6.14\!\times\!10^{-15}$ & $6.14\!\times\!10^{-14}$ \\
 & QA CS & 0.000977 & 0.004885 \\
Qwen3-VL-30B-A3B & MC CF & $1.05\!\times\!10^{-5}$ & $8.39\!\times\!10^{-5}$ \\
 & MC CS & 0.030884 & 0.123535 \\
\bottomrule
\end{tabular*}
\caption{\textbf{Exact two-sided McNemar tests for SPC on the 230-pair
comparisons.} Holm adjustment covers the ten comparisons in the table.}
\label{tab:primary-holm}
\end{table}

Model generation is greedy. Each fixed setting is evaluated once because
decoding is deterministic. We report exact repair/harm counts, exact two-sided
McNemar tests, and 10,000 bootstrap resamples of image pairs with seed 20260712.
The balanced claim-order test uses 10,000 paired resamples with seed
20260715. Holm adjustment covers 32 comparisons across models, question forms,
claim orders, and CF/CS sides. Visual CounterFact also uses 10,000 resamples
with seed 20260715. Its Holm adjustment covers four model and side
comparisons. Each image pair is one bootstrap sample.

Experiments use eight 24GB RTX 3090 GPUs with Python 3.13.9, PyTorch 2.10.0,
CUDA 12.8, and Transformers 4.57.6. Median processing time per example is
1.870/3.254 seconds for Qwen QA/MC and 14.278/21.963 seconds for LLaVA QA/MC.
SPC uses one original prediction plus image-conditioned and text-only candidate
scoring. The evaluation code scores every answer under one condition in a
single batch. Predicting the correction strength and applying selective
answer revision require no additional model pass.

\bibliography{references}

@article{chen2026cdh,
  author = {Chen, Kesheng and Hu, Yamin and Zhou, Qi and Zhu, Zhenqian and Luo, Wenjian},
  title = {{CDH-Bench}: A Commonsense-Driven Hallucination Benchmark for Evaluating Visual Fidelity in Vision-Language Models},
  journal = {arXiv preprint arXiv:2603.27982},
  year = {2026},
  eprint = {2603.27982},
  archivePrefix = {arXiv},
  primaryClass = {cs.CV},
  doi = {10.48550/arXiv.2603.27982},
  url = {https://arxiv.org/abs/2603.27982}
}

@inproceedings{rohrbach2018chair,
  author = {Rohrbach, Anna and Hendricks, Lisa Anne and Burns, Kaylee and Darrell, Trevor and Saenko, Kate},
  title = {Object Hallucination in Image Captioning},
  booktitle = {Proceedings of the 2018 Conference on Empirical Methods in Natural Language Processing},
  pages = {4035--4045},
  year = {2018},
  publisher = {Association for Computational Linguistics},
  doi = {10.18653/v1/D18-1437}
}

@inproceedings{li2023pope,
  author = {Li, Yifan and Du, Yifan and Zhou, Kun and Wang, Jinpeng and Zhao, Wayne Xin and Wen, Ji-Rong},
  title = {Evaluating Object Hallucination in Large Vision-Language Models},
  booktitle = {Proceedings of the 2023 Conference on Empirical Methods in Natural Language Processing},
  pages = {292--305},
  year = {2023},
  publisher = {Association for Computational Linguistics},
  doi = {10.18653/v1/2023.emnlp-main.20}
}

@inproceedings{guan2024hallusionbench,
  author = {Guan, Tianrui and Liu, Fuxiao and Wu, Xiyang and Xian, Ruiqi and Li, Zongxia and Liu, Xiaoyu and Wang, Xijun and Chen, Lichang and Huang, Furong and Yacoob, Yaser and Manocha, Dinesh and Zhou, Tianyi},
  title = {{HallusionBench}: An Advanced Diagnostic Suite for Entangled Language Hallucination and Visual Illusion in Large Vision-Language Models},
  booktitle = {Proceedings of the IEEE/CVF Conference on Computer Vision and Pattern Recognition},
  pages = {14375--14385},
  year = {2024}
}

@inproceedings{agrawal2018priors,
  author = {Agrawal, Aishwarya and Batra, Dhruv and Parikh, Devi and Kembhavi, Aniruddha},
  title = {Don't Just Assume; Look and Answer: Overcoming Priors for Visual Question Answering},
  booktitle = {Proceedings of the IEEE Conference on Computer Vision and Pattern Recognition},
  pages = {4971--4980},
  year = {2018}
}

@inproceedings{goyal2017vqa,
  author = {Goyal, Yash and Khot, Tejas and Summers-Stay, Douglas and Batra, Dhruv and Parikh, Devi},
  title = {Making the {V} in {VQA} Matter: Elevating the Role of Image Understanding in Visual Question Answering},
  booktitle = {Proceedings of the IEEE Conference on Computer Vision and Pattern Recognition},
  pages = {6325--6334},
  year = {2017},
  doi = {10.1109/CVPR.2017.670}
}

@inproceedings{cadene2019rubi,
  author = {Cad{\`e}ne, R{\'e}mi and Dancette, Corentin and Ben-Younes, H{\'e}di and Cord, Matthieu and Parikh, Devi},
  title = {{RUBi}: Reducing Unimodal Biases for Visual Question Answering},
  booktitle = {Advances in Neural Information Processing Systems},
  volume = {32},
  pages = {839--850},
  year = {2019}
}

@inproceedings{selvaraju2019hint,
  author = {Selvaraju, Ramprasaath R. and Lee, Stefan and Shen, Yilin and Jin, Hongxia and Ghosh, Shalini and Heck, Larry and Batra, Dhruv and Parikh, Devi},
  title = {Taking a {HINT}: Leveraging Explanations to Make Vision and Language Models More Grounded},
  booktitle = {Proceedings of the IEEE/CVF International Conference on Computer Vision},
  pages = {2591--2600},
  year = {2019}
}

@inproceedings{chen2020css,
  author = {Chen, Long and Yan, Xin and Xiao, Jun and Zhang, Hanwang and Pu, Shiliang and Zhuang, Yueting},
  title = {Counterfactual Samples Synthesizing for Robust Visual Question Answering},
  booktitle = {Proceedings of the IEEE/CVF Conference on Computer Vision and Pattern Recognition},
  pages = {10800--10809},
  year = {2020}
}

@inproceedings{niu2021cfvqa,
  author = {Niu, Yulei and Tang, Kaihua and Zhang, Hanwang and Lu, Zhiwu and Hua, Xian-Sheng and Wen, Ji-Rong},
  title = {Counterfactual {VQA}: A Cause-Effect Look at Language Bias},
  booktitle = {Proceedings of the IEEE/CVF Conference on Computer Vision and Pattern Recognition},
  pages = {12700--12710},
  year = {2021}
}

@article{wang2023amber,
  author = {Wang, Junyang and Wang, Yuhang and Xu, Guohai and Zhang, Jing and Gu, Yukai and Jia, Haitao and Wang, Jiaqi and Xu, Haiyang and Yan, Ming and Zhang, Ji and Sang, Jitao},
  title = {{AMBER}: An {LLM}-Free Multi-Dimensional Benchmark for {MLLMs} Hallucination Evaluation},
  journal = {arXiv preprint arXiv:2311.07397},
  year = {2023}
}

@inproceedings{sun2024mmhal,
  author = {Sun, Zhiqing and Shen, Sheng and Cao, Shengcao and Liu, Haotian and Li, Chunyuan and Shen, Yikang and Gan, Chuang and Gui, Liangyan and Wang, Yu-Xiong and Yang, Yiming and Keutzer, Kurt and Darrell, Trevor},
  title = {Aligning Large Multimodal Models with Factually Augmented {RLHF}},
  booktitle = {Findings of the Association for Computational Linguistics: ACL 2024},
  pages = {13088--13110},
  year = {2024},
  publisher = {Association for Computational Linguistics},
  doi = {10.18653/v1/2024.findings-acl.775}
}

@inproceedings{leng2024vcd,
  author = {Leng, Sicong and Zhang, Hang and Chen, Guanzheng and Li, Xin and Lu, Shijian and Miao, Chunyan and Bing, Lidong},
  title = {Mitigating Object Hallucinations in Large Vision-Language Models through Visual Contrastive Decoding},
  booktitle = {Proceedings of the IEEE/CVF Conference on Computer Vision and Pattern Recognition},
  pages = {13872--13882},
  year = {2024}
}

@inproceedings{favero2024m3id,
  author = {Favero, Alessandro and Zancato, Luca and Trager, Matthew and Choudhary, Siddharth and Perera, Pramuditha and Achille, Alessandro and Swaminathan, Ashwin and Soatto, Stefano},
  title = {Multi-Modal Hallucination Control by Visual Information Grounding},
  booktitle = {Proceedings of the IEEE/CVF Conference on Computer Vision and Pattern Recognition},
  pages = {14303--14312},
  year = {2024},
  doi = {10.1109/CVPR52733.2024.01356}
}

@inproceedings{huang2024opera,
  author = {Huang, Qidong and Dong, Xiaoyi and Zhang, Pan and Wang, Bin and He, Conghui and Wang, Jiaqi and Lin, Dahua and Zhang, Weiming and Yu, Nenghai},
  title = {{OPERA}: Alleviating Hallucination in Multi-Modal Large Language Models via Over-Trust Penalty and Retrospection-Allocation},
  booktitle = {Proceedings of the IEEE/CVF Conference on Computer Vision and Pattern Recognition},
  pages = {13418--13427},
  year = {2024}
}

@inproceedings{chen2024halc,
  author = {Chen, Zhaorun and Zhao, Zhuokai and Luo, Hongyin and Yao, Huaxiu and Li, Bo and Zhou, Jiawei},
  title = {{HALC}: Object Hallucination Reduction via Adaptive Focal-Contrast Decoding},
  booktitle = {Proceedings of the 41st International Conference on Machine Learning},
  series = {Proceedings of Machine Learning Research},
  volume = {235},
  pages = {7824--7846},
  year = {2024},
  publisher = {PMLR}
}

@inproceedings{liu2024pai,
  author = {Liu, Shi and Zheng, Kecheng and Chen, Wei},
  title = {Paying More Attention to Image: A Training-Free Method for Alleviating Hallucination in {LVLMs}},
  booktitle = {Computer Vision -- ECCV 2024},
  pages = {125--140},
  year = {2024},
  publisher = {Springer},
  doi = {10.1007/978-3-031-73010-8_8}
}

@inproceedings{liu2025mfcd,
  author = {Liu, Bingqian and Zhang, Fu and Chen, Guoqing and Cheng, Jingwei},
  title = {Multi-Frequency Contrastive Decoding: Alleviating Hallucinations for Large Vision-Language Models},
  booktitle = {Proceedings of the 2025 Conference on Empirical Methods in Natural Language Processing},
  pages = {28568--28584},
  year = {2025},
  publisher = {Association for Computational Linguistics},
  doi = {10.18653/v1/2025.emnlp-main.1452}
}

@article{wu2026revis,
  author = {Wu, Jialin and Shi, Wei and Shen, Han and Qi, Peigui and Tang, Kunsheng and Huang, Zhicong and Wang, Binghao and Yang, Zhou},
  title = {Revis: Sparse Latent Steering to Mitigate Object Hallucination in Large Vision-Language Models},
  journal = {arXiv preprint arXiv:2602.11824},
  year = {2026},
  note = {Accepted at ICML 2026}
}

@article{ren2026nolan,
  author = {Ren, Lingfeng and Yu, Weihao and Yu, Runpeng and Wang, Xinchao},
  title = {{NoLan}: Mitigating Object Hallucinations in Large Vision-Language Models via Dynamic Suppression of Language Priors},
  journal = {arXiv preprint arXiv:2602.22144},
  year = {2026},
  eprint = {2602.22144},
  archivePrefix = {arXiv},
  primaryClass = {cs.CV},
  doi = {10.48550/arXiv.2602.22144},
  url = {https://arxiv.org/abs/2602.22144}
}

@inproceedings{zhou2023rome,
  author = {Zhou, Kankan and Lai, Eason and Yeong, Wei Bin Au and Mouratidis, Kyriakos and Jiang, Jing},
  title = {{ROME}: Evaluating Pre-trained Vision-Language Models on Reasoning beyond Visual Common Sense},
  booktitle = {Findings of the Association for Computational Linguistics: EMNLP 2023},
  pages = {10185--10197},
  year = {2023},
  publisher = {Association for Computational Linguistics},
  doi = {10.18653/v1/2023.findings-emnlp.683}
}

@inproceedings{lee2025vlind,
  author = {Lee, Kang-il and Kim, Minbeom and Yoon, Seunghyun and Kim, Minsung and Lee, Dongryeol and Koh, Hyukhun and Jung, Kyomin},
  title = {{VLind-Bench}: Measuring Language Priors in Large Vision-Language Models},
  booktitle = {Findings of the Association for Computational Linguistics: NAACL 2025},
  pages = {4129--4144},
  year = {2025},
  publisher = {Association for Computational Linguistics},
  doi = {10.18653/v1/2025.findings-naacl.231}
}

@inproceedings{liu2025insight,
  author = {Liu, Xiaoyuan and Wang, Wenxuan and Yuan, Youliang and Huang, Jen-tse and Liu, Qiuzhi and He, Pinjia and Tu, Zhaopeng},
  title = {Insight Over Sight: Exploring the Vision-Knowledge Conflicts in Multimodal {LLM}s},
  booktitle = {Proceedings of the 63rd Annual Meeting of the Association for Computational Linguistics},
  pages = {17825--17846},
  year = {2025},
  publisher = {Association for Computational Linguistics},
  doi = {10.18653/v1/2025.acl-long.872}
}

@inproceedings{zhou2025causalmm,
  author = {Zhou, Guanyu and Yan, Yibo and Zou, Xin and Wang, Kun and Liu, Aiwei and Hu, Xuming},
  title = {Mitigating Modality Prior-Induced Hallucinations in Multimodal Large Language Models via Deciphering Attention Causality},
  booktitle = {The Thirteenth International Conference on Learning Representations},
  year = {2025}
}

@inproceedings{golovanevsky2025pixels,
  author = {Golovanevsky, Michal and Rudman, William and Lepori, Michael and Bar, Amir and Singh, Ritambhara and Eickhoff, Carsten},
  title = {Pixels Versus Priors: Controlling Knowledge Priors in Vision-Language Models through Visual Counterfacts},
  booktitle = {Proceedings of the 2025 Conference on Empirical Methods in Natural Language Processing},
  pages = {24837--24852},
  year = {2025},
  publisher = {Association for Computational Linguistics},
  doi = {10.18653/v1/2025.emnlp-main.1262}
}

@inproceedings{zheng2025reefknot,
  author = {Zheng, Kening and Chen, Junkai and Yan, Yibo and Zou, Xin and Zhou, Huiyu and Hu, Xuming},
  title = {Reefknot: A Comprehensive Benchmark for Relation Hallucination Evaluation, Analysis and Mitigation in Multimodal Large Language Models},
  booktitle = {Findings of the Association for Computational Linguistics: ACL 2025},
  pages = {6193--6212},
  year = {2025},
  publisher = {Association for Computational Linguistics},
  doi = {10.18653/v1/2025.findings-acl.322}
}

@article{bai2025qwen3vl,
  author = {Bai, Shuai and Cai, Yuxuan and Chen, Ruizhe and Chen, Keqin and Chen, Xionghui and Cheng, Zesen and Deng, Lianghao and Ding, Wei and Gao, Chang and Ge, Chunjiang and others},
  title = {{Qwen3-VL} Technical Report},
  journal = {arXiv preprint arXiv:2511.21631},
  year = {2025}
}

@inproceedings{liu2024llavabaselines,
  author = {Liu, Haotian and Li, Chunyuan and Li, Yuheng and Lee, Yong Jae},
  title = {Improved Baselines with Visual Instruction Tuning},
  booktitle = {Proceedings of the IEEE/CVF Conference on Computer Vision and Pattern Recognition},
  pages = {26296--26306},
  year = {2024}
}

@inproceedings{li2026popev2,
  author = {Li, Yifan and Zhou, Kun and Zhao, Wayne Xin and Fang, Lei and Wen, Ji-Rong},
  title = {Analyzing and Mitigating Object Hallucination: A Training Bias Perspective},
  booktitle = {Proceedings of the AAAI Conference on Artificial Intelligence},
  volume = {40},
  pages = {6636--6643},
  year = {2026},
  doi = {10.1609/aaai.v40i8.37594},
  url = {https://ojs.aaai.org/index.php/AAAI/article/view/37594}
}

\end{document}